# Small Language Models Offer Significant Potential for Science Community

**Jian Zhang (张健)[1*]**

[1] School of Geophysics and Geomatics, China University of Geosciences, Wuhan 430074, China

*Correspondence to: Jian Zhang, zhangjian@cug.edu.cn

**Abstract** Recent advancements in natural language processing, particularly with large language models (LLMs), are transforming how scientists engage with the literature. While the adoption of LLMs is increasing, concerns remain regarding potential information biases and computational costs. Rather than LLMs, I developed a framework to evaluate the feasibility of precise, rapid, and cost-effective information retrieval from extensive geoscience literature using freely available small language models (MiniLMs). A curated corpus of approximately 77 million high-quality sentences, extracted from 95 leading peer-reviewed geoscience journals such as *Geophysical Research Letters* and *Earth and Planetary Science Letters* published during years 2000 to 2024, was constructed. MiniLMs enable a computationally efficient approach for extracting relevant domain-specific information from these corpora through semantic search techniques and sentence-level indexing. This approach, unlike LLMs such as ChatGPT-4 that often produces generalized responses, excels at identifying substantial amounts of expert-verified information with established, multi-disciplinary sources, especially for information with quantitative findings. Furthermore, by analyzing emotional tone via sentiment analysis and topical clusters through unsupervised clustering within sentences, MiniLM provides a powerful tool for tracking the evolution of conclusions, research priorities, advancements, and emerging questions within geoscience communities. Overall, MiniLM holds significant potential within the geoscience community for applications such as fact and image retrievals, trend analyses, contradiction analyses, and educational purposes.

**Key points:**

- A framework integrating small language models with an extensive and high-quality corpus of sentences from geoscience journals was established
- This framework facilitates precise and cost-effective knowledge transfer within geosciences, improving the reliability of information extraction
- Analyzing emotional tone and topical clusters within sentences aids in understanding perspectives on critical issues such as climate change

**Plain Language Summary** The field of geoscience (or any other field) has experienced an unprecedented burst in research publications, reflecting a rapidly advancing understanding of our planet. This expansion has resulted in an increasingly complex body of knowledge and a cognitive overload, while a concurrent shift towards interdisciplinary and transdisciplinary research poses challenges to effective and cheap knowledge transfer. To address these challenges, this perspective introduces an approach that leverages the extensive body of Earth science literature. By analyzing millions of individual sentences extracted from literature using small language models, the method enables a cost-effective and precise means of acquiring professional knowledge. Integrating semantic analysis with large language model summarization, the method has the potential to yield more specialized and quantitative insights than current excellent models like ChatGPT-4.

# 1 Introduction

The field of Earth science is undergoing rapid evolution in the Anthropocene, largely driven by the integration of artificial intelligence (AI) technologies. This necessitates robust interdisciplinary collaborations between computer science, data science, and geoscience (Bergen *et al*., 2019; Cui, 2021; Li *et al*., 2023). Such collaborations are increasingly vital given the pervasive impacts of climate change (Nielsen & D"haen, 2014). The Earth is a complex, interconnected, and dynamic system (Vance *et al*., 2024), characterized by intricate human-environment relationships, complicated physical, dynamic, and chemical interactions, and variations across multiple temporal and spatial scales. For instance, river basins emerge from the complex interplay among land, biosphere, atmosphere, and human systems, exhibiting significant uncertainties across both temporal and spatial scales (Best, 2019). Precipitation, a vital component of the water cycle, links the hydrosphere, lithosphere, biosphere, and atmosphere (Wu *et al*., 2020). These inherent complexities present substantial challenges for researchers seeking to acquire comprehensive and nuanced understanding.

Recent advancements in natural language processing, particularly in large language model (LLM) technology, offer substantial new avenues for scientific research. These models possess the potential to transform scientific practices. For example, LLMs can significantly enhance the efficiency and scale of ecological research (Gougherty & Clipp, 2024). Furthermore, LLMs such as the Generative Pre-trained Transformer (GPT) can be readily adapted for various tasks in chemistry and materials science by fine-tuning them to provide accurate responses to scientific queries expressed in natural language (Jablonka *et al*., 2024).

However, it is crucial to acknowledge that responses generated by LLMs can sometimes be misleading, inaccurate, or factually incorrect (Foroumandi *et al*., 2023). Therefore, validating information derived from LLMs against established sources, such as peer-reviewed publications and scholarly books, is essential (Foroumandi *et al*., 2023). Furthermore, the black-box nature of LLMs often obscures the reasoning behind their predictions (Ott *et al*., 2023). Additionally, the current limitations in accessing external knowledge sources can restrict the utility of LLMs in scientific applications (M. Bran *et al*., 2024). As Li *et al*. (2023) noted, "While foundation models have had great success in applications such as natural language processing, how or if they could be applied in geosciences remains to be seen."

Peer-reviewed publications constitute the primary repository of high-quality scientific knowledge, often containing latent insights relevant to future discoveries (Tshitoyan *et al*., 2019). However, the unstructured nature and substantial heterogeneity within these publications present a significant obstacle to large-scale information analysis (Kononova *et al*., 2021).

Corpus is the heart of a language model. Authoritative and curated sources is crucial for the creation of LLM applications for science (Ramachandran & Bugbee, 2025). Leading journals, such as *Geophysical Research Letters*, *Earth and Planetary Science Letters*, and *Nature Geoscience*, represent primary sources for constructing high-quality corpora in the geosciences. These publications encapsulate the collective expertise and insights of numerous researchers, reflecting a substantial accumulation of knowledge spanning decades. By developing a comprehensive corpus from these authoritative sources and leveraging high-performance language models, research capabilities could be significantly augmented, potentially facilitating the generation of novel hypotheses. It is noteworthy that a language model does not necessarily need to be large; large models may sometimes limit their applicability and interpretability.

## 2 Material and Method

### 2.1 Journal sources

The data is derived from a broad range of geoscience journals published by major academic entities. These include the Nature portfolio, Elsevier, Wiley, Springer, the American Association for the Advancement of Science (AAAS), the American Meteorological Society (AMS), the American Geophysical Union (AGU), and the European Geosciences Union (EGU). Collectively, these sources provide a robust and rigorous body of peer-reviewed geoscience research.

### 2.2 Standardized literature

From these journal sources, approximately 95 highly reputable geoscience journals were rigorously selected in this perspective. This selection includes prominent publications such as *Nature Geoscience*, *Geophysical Research Letters*, *Journal of Geophysical Research*, *Remote Sensing of Environment*, *Earth-Science Reviews*, *Science Advances*, *Earth and Planetary Science Letters*, *Journal of the Atmospheric Sciences*, *Geochimica et Cosmochimica Acta*, *Environmental Science & Technology*, *Geology*, *Proceedings of the National Academy of Sciences*, and *Water Research*. These journals represent a broad spectrum of geoscience disciplines, including atmospheric science, space science, hydrology, environmental science, remote sensing, oceanography, ecology, and geophysics.

Consequently, I compiled a dataset of over 376,000 articles published between 2000 and 2024. To ensure consistency, Portable Document Format (PDF) filenames adhere to a standardized format: "journal abbreviation–year of publication–title." A comprehensive list of journal names, their corresponding abbreviations, and the total article count for each journal is available in Table 1.

**Table 1.** *Journal and sentence information. This table delineates comprehensive information about the journal, encompassing its full name, International Organization for Standardization (ISO) abbreviation (only the Latin letters are retained), years of literature coverage, total number of published articles, and the aggregate literature sentences.*

| Full name of journal | ISO Abbreviation | Years | Article no. | Sentence no. |
|---|---|---|---|---|
| *Geophysical Research Letters* | ***GeophysResLett*** | 2002–2024 | 31,820 | 3,305,108 |
| *Earth's Future* | ***EarthsFuture*** | 2014–2024 | 1,479 | 344,903 |
| *Reviews of Geophysics* | ***RevGeophys*** | 2002–2024 | 443 | 174,007 |
| *Water Resources Research* | ***WaterResourRes*** | 2002–2024 | 11,669 | 2,597,772 |
| *Journal of Geophysical Research: Atmospheres* | ***JGeophysResAtmos*** | 2002–2024 | 17,458 | 3,953,872 |
| *Journal of Geophysical Research: Oceans* | ***JGeophysResOceans*** | 2002–2024 | 9,206 | 2,089,057 |
| *Journal of Geophysical Research: Planets* | ***JGeophysResPlanets*** | 2002–2024 | 3,770 | 890,708 |
| *Journal of Geophysical Research: Space Physics* | ***JGeophysResSpacePhys*** | 2002–2024 | 13,907 | 2,741,795 |
| *Journal of Geophysical Research: Solid Earth* | ***JGeophysResSolidEarth*** | 2002–2024 | 10,628 | 2,471,150 |
| *Journal of Geophysical Research: Biogeosciences* | ***JGeophysResBiogeosci*** | 2005–2024 | 3,524 | 749,651 |
| *Journal of Geophysical Research: Earth Surface* | ***JGeophysResEarthSurf*** | 2003–2024 | 2,778 | 693,676 |
| *Space Weather* | ***SpaceWeather*** | 2003–2024 | 1,549 | 296,054 |
| *AGU Advances* | ***AGUAdv*** | 2020–2024 | 200 | 47,092 |
| *Journal of Geophysical Research: Machine Learning and Computation* | ***JGeophysResMachLearnComput*** | 2024–2024 | 62 | 17,741 |
| *Journal of Advances in Modeling Earth Systems* | ***JAdvModelEarthSyst*** | 2009–2024 | 2,017 | 508,536 |
| *Perspectives of Earth and Space Scientists* | ***PerspectEarthSpaceSci*** | 2023–2024 | 16 | 2,947 |
| *Geochemistry Geophysics Geosystems* | ***GeochemGeophysGeosyst*** | 2002–2024 | 5,300 | 1,069,332 |
| *Radio Science* | ***RadioSci*** | 2023–2024 | 69 | 12,616 |
| *Global Biogeochemical Cycles* | ***GlobBiogeochemCycle*** | 2002–2024 | 2,530 | 535,221 |
| *Earth and Space Science* | ***EarthSpaceSci*** | 2014–2024 | 1,610 | 332,285 |
| *Tectonics* | ***Tectonics*** | 2023–2024 | 118 | 34,688 |
| *Global Environmental Change-human And Policy Dimensions* | ***GlobEnvironChangeHumanPolicyDimens*** | 2010–2024 | 1,895 | 476,964 |
| *Earth and Planetary Science Letters* | ***EarthPlanetSciLett*** | 2004–2024 | 10,001 | 1,867,375 |
| *Earth-Science Reviews* | ***EarthSciRev*** | 2009–2024 | 2,706 | 1,020,985 |
| *ISPRS Journal of Photogrammetry and Remote Sensing* | ***ISPRSJPhotogrammRemoteSens*** | 2000–2024 | 3,067 | 812,215 |
| *Journal of Hydrology* | ***JHydrol*** | 2009–2024 | 14,280 | 3,360,058 |
| *Remote Sensing of Environment* | ***RemoteSensEnviron*** | 2003–2024 | 7,596 | 1,924,737 |
| *Water Research* | ***WaterRes*** | 2003–2024 | 15,161 | 2,904,672 |
| *Weather and Climate Extremes* | ***WeatherClimExtremes*** | 2013–2024 | 634 | 138,841 |
| *Agricultural and Forest Meteorology* | ***AgricForMeteorol*** | 2009–2024 | 4,208 | 921,866 |
| *Geochimica et Cosmochimica Acta* | ***GeochimCosmochimActa*** | 2009–2024 | 6,832 | 1,783,798 |
| *Advances in Water Resources* | ***AdvWaterResour*** | 2009–2024 | 2,781 | 702,577 |
| *One Earth* | ***OneEarth*** | 2019–2024 | 435 | 50,131 |
| *Science Bulletin* | ***SciBull*** | 2016–2024 | 3,098 | 371,702 |
| *Lancet Planetary Health* | ***LancetPlanetHealth*** | 2017–2024 | 387 | 51,932 |
| *Renewable and Sustainable Energy Reviews* | ***RenewSustEnergRev*** | 2009–2024 | 12,850 | 3,548,197 |
| *Environment International* | ***EnvironInt*** | 2009–2024 | 6,576 | 1,307,122 |
| *Applied Geography* | ***ApplGeogr*** | 2009–2024 | 2,202 | 482,631 |
| *Cities* | ***Cities*** | 2010–2024 | 3,068 | 772,181 |
| *Computers, Environment and Urban Systems* | ***ComputEnvironUrbanSyst*** | 2009–2021 | 817 | 205,200 |
| *Journal of Hazardous Materials* | ***JHazardMater*** | 2009–2024 | 14,592 | 2,690,139 |
| *Geoderma* | ***Geoderma*** | 2009–2024 | 4,847 | 979,260 |
| *International Journal of Applied Earth Observation and Geoinformation* | ***IntJApplEarthObsGeoinf*** | 2009–2024 | 2,773 | 599,791 |
| *Journal of Rural Studies* | ***JRuralStud*** | 2010–2024 | 1,995 | 565,611 |
| *Landscape and Urban Planning* | ***LandscUrbanPlan*** | 2009–2024 | 2,326 | 495,518 |
| *Land Use Policy* | ***LandUsePol*** | 2009–2024 | 4,980 | 1,231,936 |
| *Nature Geoscience* | ***NatGeosci*** | 2007–2024 | 2,014 | 182,328 |
| *Communications Earth & Environment* | ***CommunEarthEnviron*** | 2020–2024 | 1,085 | 170,353 |
| *Nature Ecology & Evolution* | ***NatEcolEvol*** | 2016–2024 | 736 | 80,675 |
| *Nature Reviews Earth & Environment* | ***NatRevEarthEnviron*** | 2019–2024 | 91 | 20,412 |
| *Nature Climate Change* | ***NatClimChang*** | 2011–2024 | 1,158 | 108,836 |
| *Nature Water* | ***NatWater*** | 2023–2024 | 41 | 8,395 |
| *Nature Communications (Geoscience only)* | ***NatCommun*** | 2014–2024 | 4,961 | 651,741 |
| *npj Climate and Atmospheric Science* | ***npjClimAtmosSci*** | 2018–2024 | 519 | 78,739 |
| *Nature (Geoscience only)* | ***Nature*** | 2009–2024 | 1,449 | 164,475 |
| *Nature Sustainability* | ***NatSustain*** | 2018–2024 | 434 | 56,801 |
| *Nature Energy* | ***NatEnergy*** | 2015–2024 | 406 | 55,454 |
| *Atmospheric Chemistry and Physics* | ***AtmosChemPhys*** | 2001–2024 | 12,461 | 3,298,928 |
| *Earth System Science Data* | ***EarthSystSciData*** | 2009–2024 | 1,200 | 269,809 |
| *Atmospheric Measurement Techniques* | ***AtmosMeasTech*** | 2008–2024 | 4,367 | 1,124,558 |
| *Hydrology and Earth System Sciences* | ***HydrolEarthSystSci*** | 2001–2024 | 4,872 | 1,222,152 |
| *Journal of Applied Meteorology and Climatology* | ***JApplMeteorolClimatol*** | 2010–2024 | 2,072 | 473,546 |
| *Journal of the Atmospheric Sciences* | ***JAtmosSci*** | 2003–2024 | 4,196 | 1,070,337 |
| *Journal of Climate* | ***JClim*** | 2000–2024 | 9,364 | 2,146,492 |
| *Monthly Weather Review* | ***MonWeatherRev*** | 2005–2024 | 4,343 | 1,083,666 |
| *Journal of Atmospheric and Oceanic Technology* | ***JAtmosOceanTechnol*** | 2005–2024 | 2,806 | 612,599 |
| *National Science Review (Geoscience only)* | ***NatlSciRev*** | 2014–2020 | 105 | 11,799 |
| *Geology* | ***Geology*** | 2003–2020 | 3,756 | 81,821 |
| *Science (Geoscience only)* | ***Science*** | 2014–2024 | 415 | 21,153 |
| *Science Advances (Geoscience only)* | ***SciAdv*** | 2015–2024 | 2,147 | 398,591 |
| *Environmental Science & Technology* | ***EnvironSciTechnol*** | 2010–2024 | 21,060 | 3,024,293 |
| *Environmental Research Letters* | ***EnvironResLett*** | 2010–2020 | 2,264 | 343,357 |
| *Global Change Biology* | ***GlobChangeBiol*** | 2013–2024 | 3,209 | 440,763 |
| *Ecology Letters* | ***EcolLett*** | 2013–2024 | 1,973 | 352,645 |
| *Ecological Monographs* | ***EcolMonogr*** | 2010–2024 | 492 | 157,715 |
| *Land Degradation & Development* | ***LandDegradDev*** | 2010–2024 | 3,061 | 570,632 |
| *Global Ecology and Biogeography* | ***GlobEcolBiogeogr*** | 2013–2024 | 151 | 26,350 |
| *Wiley Interdisciplinary Reviews-Climate Change* | ***WileyInterdiscipRevClimChang*** | 2013–2024 | 330 | 73,138 |
| *Wiley Interdisciplinary Reviews-Water* | ***WileyInterdiscipRevWater*** | 2018–2024 | 338 | 77,307 |
| *IEEE Geoscience and Remote Sensing Magazine* | ***IEEEGeosciRemoteSensMag*** | 2015–2024 | 288 | 58,936 |
| *IEEE Journal of Selected Topics in Applied Earth Observations and Remote Sensing* | ***IEEEJSelTopApplEarthObservRemoteSens*** | 2015–2024 | 3,400 | 747,643 |
| *IEEE Transactions on Geoscience and Remote Sensing* | ***IEEETransGeosciRemoteSensing*** | 2001–2024 | 12,489 | 2,950,842 |
| *Science China- Earth Sciences* | ***SciChinaEarthSci*** | 2016–2024 | 1,073 | 215,832 |
| *Current Climate Change Reports* | ***CurrClimChangRep*** | 2015–2024 | 131 | 24,846 |
| *Climatic Change* | ***ClimChange*** | 2010–2024 | 2,871 | 478,813 |
| *Mathematical Geosciences* | ***MathGeosci*** | 2010–2024 | 588 | 108,302 |
| *Journal of Geodesy* | ***JGeodesy*** | 2010–2024 | 938 | 209,072 |
| *International Journal of Geographical Information Science* | ***IntJGeogrInfSci*** | 2010–2023 | 1,357 | 316,206 |
| *Urban Geography* | ***UrbanGeogr*** | 2009–2023 | 953 | 207,811 |
| *Annals of the American Association of Geographers* | ***AnnAmAssocGeogr*** | 2009–2024 | 1,158 | 236,741 |
| *Critical Reviews in Environmental Science and Technology* | ***CritRevEnvironSciTechnol*** | 2010–2021 | 566 | 183,037 |
| *Annual Review of Environment and Resources* | ***AnnuRevEnvironResour*** | 2005–2024 | 279 | 89,135 |
| *Annual Review of Marine Science* | ***AnnuRevMarSci*** | 2012–2024 | 240 | 64,239 |
| *Annual Review of Earth and Planetary Sciences* | ***AnnuRevEarthPlanetSci*** | 2007–2024 | 478 | 152,760 |
| *Proceedings of the National Academy of Sciences of the United States of America (Geoscience only)* | ***ProcNatlAcadSciUSA*** | 2007–2024 | 2,036 | 225,256 |

## 2.3 Body text identification and sentence splitting

Due to the intricate formatting of PDF documents, which incorporates elements like icons, publisher and article details, figures, tables, and their associated captions, accurate extraction of body text is challenging. I addressed this issue by focusing on the extraction of the "Abstract" and main body text, while deliberately excluding sections such as the "Reference". For articles with two- or three-column layouts, text segmentation was implemented on each page to mimic human reading habit. Sentence segmentation relied on the period character ("."), with exceptions made for common abbreviations such as "et al.", "Fig.", and "Tab.". Sentences longer than 256 or shorter than 10 words were discarded. The final processed dataset comprised 76,862,981 extracted sentences (Tab. 1).

**Table 2.** *List of pre-trained sentence transformer models. This table delineates the comprehensive nomenclature, abbreviated designation, maximum sequence length, vector dimensionality, and overall model size of the respective models, where "M" denotes million and "B" signifies billion.*

| Full name of Model | abbr. | Max Sequence Length | Dimension | Model Size |
|---|---|---|---|---|
| *all-MiniLM-L6-v2* | **PSTM_1** | 256 | 384 | 22.7 M |
| *all-MiniLM-L12-v2* | **PSTM_2** | 256 | 384 | 33.4 M |
| *all-mpnet-base-v2* | **PSTM_3** | 384 | 768 | 109 M |
| *mxbai-embed-large-v1* | **PSTM_4** | 512 | 1,024 | 335 M |
| *multilingual-e5-large-instruct* | **PSTM_5** | 512 | 1,024 | 560 M |
| *SFR-Embedding-Mistra* | **PSTM_6** | 4,096 | 4,096 | 7.11 B |

## 2.4 Topic selection and word embedding

To minimize computational requirements, a keyword-based approach was initially employed to identify and match sentences across the entire corpus. For instance, in studies focused on precipitation, keywords such as "precipitation" (case-insensitive) and " rain" (case-insensitive) were utilized. These filtered sentences were then encoded into dense vectors using six pre-trained sentence transformer models (PSTMs), which have demonstrated high performance in the Massive Text Embedding Benchmark (MTEB). The selected PSTMs encompass both lightweight models, including *all-MiniLM-L6-v2* and *all-MiniLM-L12-v2* (Reimers & Gurevych, 2019a), designed for a balance of speed and performance, and medium-to-large models like *SFR-Embedding-Mistra* (Meng *et al.*, 2024). The parameter counts of these models range significantly, from 22.7 million to 7.11 billion, with dense vector embeddings having dimensionalities ranging from 384 to 4096. The maximum input token sequence length varies from 256 to 4096 tokens (detailed parameters are available in Table 2). It should be noted that top-performing models on the MTEB benchmark do not necessarily transfer effectively to specific tasks (Reimers & Gurevych, 2019b).

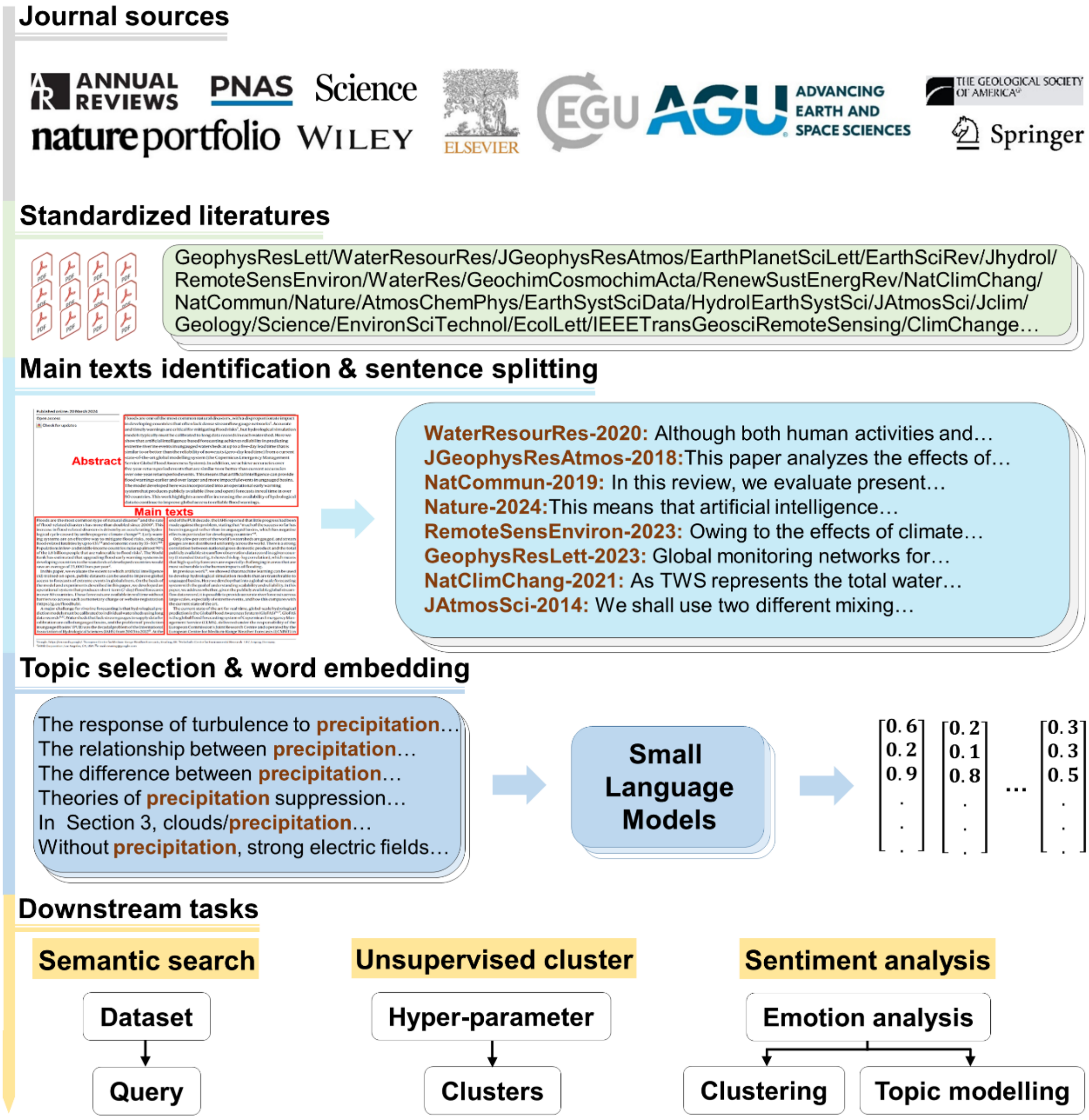

**Figure 1.** The concept diagram of sentence-level MiniLM approaches for geosciences. The methodology follows a hierarchical framework: initially, literature sources are compiled and standardized (refer to Tab. 1 for journal abbreviations). Subsequently, body texts are identified and segmented into individual sentences (see Tabs. 1 total sentence counts). Following this, topics are selected, and word embeddings are generated (see Tab. 2 for details on PSTMs). Finally, three downstream tasks are executed: Task 1 (semantic search) is illustrated in Figs. 2–3 and Figs. S1–S11; Task 2 (clustering) is represented in Figs. 4–5; and Task 3 (sentiment analysis) is shown in Figs. 6–7 and Tabs. S1–S2.

## 2.5 Downstream tasks

### 2.5.1 Sentence-level semantic search

For this task, PSTMs were utilized to encode both the text corpora and the user-submitted query into dense vector representations. Similarity scores between the corpora and query were calculated by taking the inner product of their respective vectors, where higher scores signify greater relevance. For the ensemble model, a final relevance ranking was produced by averaging the similarity scores derived from the six individual PSTMs. The highest-ranked sentences were then processed by LLMs (Llama, DeepSeek or ChatGPT-4) to automatically summarize the main themes.

#### 2.5.2 Sentence-level unsupervised clustering

To prepare for unsupervised clustering using dense vectors obtained from a PSTM, the categorization of sentences into meaningful groups is practicable. For instance, sentences can be grouped by their year of publication. Agglomerative Clustering was selected as the clustering method because it allows for the specification of a merging threshold. This is beneficial when the actual number of clusters is unclear, allowing for the manual control of cluster granularity through modification of the threshold. In addition, hyperparameters such as a minimum cluster count and a minimum similarity score are necessarily defined.

#### 2.5.3 Sentence-level sentiment analysis

The *go-emotion* language model architecture can categorize sentences into 27 separate emotion classes, such as "admiration," "confusion," "approval," and "disappointment" (Demszky *et al*., 2020). The *Twitter-RoBERTa-base-sentiment-latest* language model can classify sentences as exhibiting negative, positive, or neutral sentiment. Sentences identified as expressing specific sentiments can be further analyzed using topic modeling (e.g., with the *BERTopic* language model) or unsupervised clustering. Figure 1 illustrates the full method workflow.

## 3 Downstream tasks

### 3.1 Sentence-level semantic search

Gaining in-depth knowledge within any field demands substantial effort. For instance, although the general impact of climate change on precipitation patterns is widely acknowledged, the nuanced details are often unavailable to those without specialized training in climatology. To investigate this relationship, all relevant sentences were first filtered. Then, PSTMs were employed to perform a semantic search and identify the most pertinent historical sources using the query: "The pattern of precipitation is modified by climate change" (Figure 2). While more complex models generally yielded higher similarity scores than a lightweight model like PSTM_1, a significant score variance of approximately 0.1 across the six models was found. This suggests that even advanced models may fail to identify the most relevant sources when used in isolation. Therefore, an ensemble model, which arithmetically averages the similarity scores of all six PSTMs, probably provides a more reliable and robust ranking system. Applying the L2 norm to assess the difference between individual PSTM scores and the average score reveals that PSTM_5 has the greatest influence on the average (29.8%), with PSTM_6 having the smallest influence (7.2%). The top-ranked sentences identified by this ensemble model effectively illustrate the complex ways in which climate change influences precipitation (Fig. 2b). Finally, a comprehensive summary was created from the top ranked sentences (5,000 in this analysis) using a pre-trained LLM (Llama 2 70B), and this summary effectively elucidates the intricate roles of climate change in modifying precipitation patterns.

While effective, keyword-based sentence filtering is limited by its inability to consider semantic context and the surrounding context. To address these limitations, I further developed a search software that can utilize all the aforementioned corpus in Tab.1. Through simple question-based queries, the software can rapidly identify the most relevant sentences and their corresponding literature sources, as illustrated in Figure 3. For example, a query such as “radiosonde has a time interval of s” submitted to the software returned recommendations based on the PTSM_3 model, along with the surrounding contextual sentences. This search yielded accurate technical specifications, indicating typical sampling frequencies of 1 or 2 seconds for modern radiosondes, with standard release time at 0000 and 1200 UTC. In contrast, ChatGPT-4 provided inaccurate results for the same query. Further direct comparisons between the software and ChatGPT-4 are presented in Figures S1–S6 in the Supporting Information. These results highlight the advantage of the software over ChatGPT-4, particularly for retrieving quantitative scientific data (e.g., turbulence dissipation rate in the lower atmosphere), navigating controversial scientific conclusions (e.g., wind changes under global change), or accessing less commonly known information (e.g., the download address of meteor radar data). The information provided by the search software is precise, comprehensive, and directly verifiable through the linked source file.

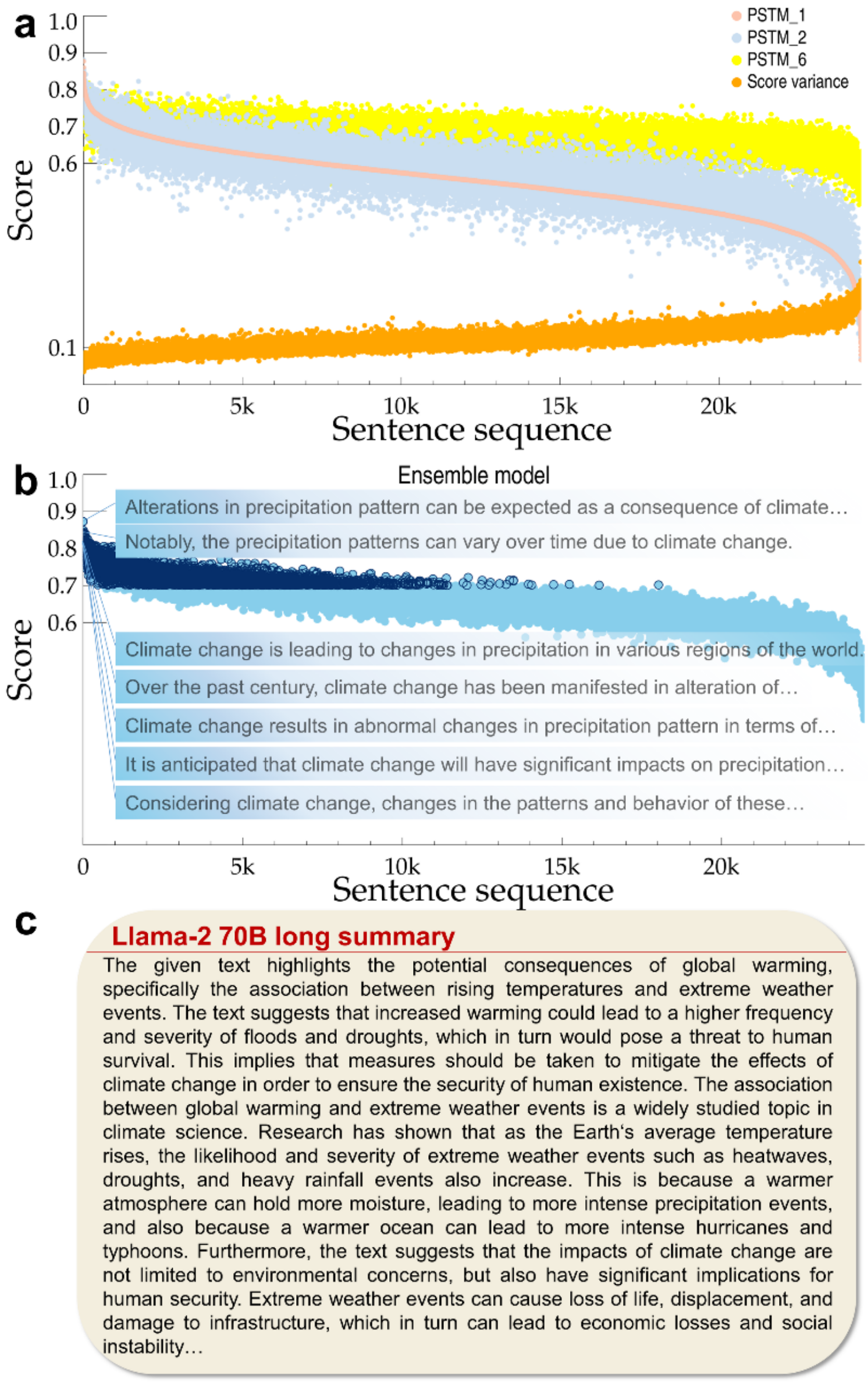

**Figure 2.** The query “The pattern of precipitation is modified by climate change” was utilized to demonstrate sentence-level semantic search. A corpus of 24,475 sentences containing “precipitation” (case-insensitive) or “ rain” (case-insensitive) and “climate change” (case-insensitive) was used for analysis. Dot product similarity scores were calculated between the query and sentence embeddings. Sentences were sorted according to PSTM_1 scores (light red), with corresponding PSTM_2 (light blue) and PSTM_6 (yellow) scores visualized in panel (a). The variance of scores across six models is indicated in orange. A comprehensive average score (panel b) was computed, and the top 5,000 ranked sentences are shown with dark blue outlines. The top 7 ranked sentences are shown in the figure. Panel (c) displays a part of summary generated using a pre-trained LLM (Llama 2 70B), prompted to “Please summarize provided text with at least 400 words”, based on the top 5,000 ranked sentences.

Moreover, the retrieval software can be effectively integrated with pre-trained LLMs. As demonstrated in Figures S7–S8, this hybrid approach, which combines semantic matching across the entire corpus with summarization of relevant sentences (similarity scores > 0.7) and their surrounding context, yields comprehensive and nuanced summaries of complex topics, such as the influence of urban heat islands on precipitation and atmospheric gravity wave formation. However, the computational demands of this integrated approach currently limit its practical application when using ensemble models, as depicted in Figure 2.

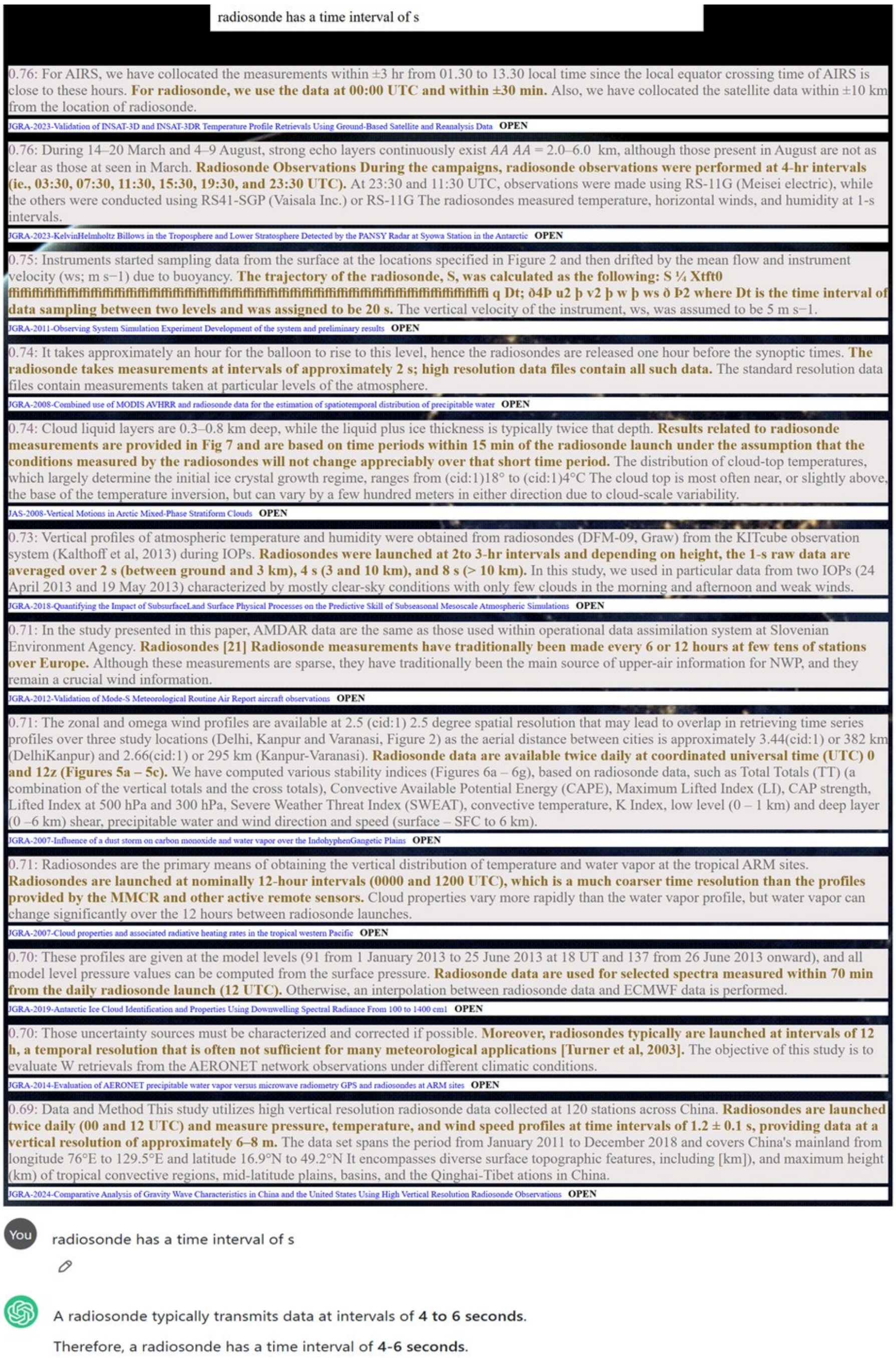

**Figure 3.** I developed a software to illustrate semantic search capabilities within atmospheric science literature. To minimize computational requirements, only atmospheric-relevant journals such as *Geophysical Research Letters* and *Journal of Geophysical Research: Atmospheres* were selected. The software was run on a laptop equipped with an Intel® Core™ i7-10750H CPU @ 2.60GHz, 64GB of memory, and an NVIDIA Quadro P620 GPU. The bottom panel replicates a query generated by ChatGPT-4: "radiosonde has a time interval of s." The top-matched sentences, identified by the PSTM_3 model, are highlighted in orange, with the corresponding relevance score indicated numerically. Contextual sentences preceding and following the matched segments are displayed in black, with blue annotations denoting the original reference. An "OPEN" button facilitates direct access to the original publication in PDF format.

Academic debates often feature a diverse array of opinions, encompassing both favorable and unfavorable perspectives on a specific subject. As an example, Figure S9 uses two contrasting queries to illustrate this point: "Machine learning and deep learning have advantages in hydrological modeling" and "Machine learning and deep learning have limitations in hydrological modeling." While these queries are apparently opposing, they show a high semantic similarity score of 0.95 when assessed by PSTM_1. This may highlight the reduced capacity of lightweight PSTMs, such as PSTM_1, to distinguish subtle semantic differences. However, medium-to-large PSTMs demonstrate enhanced performance in this domain. By mapping sentence collections into a 1,024-dimensional vector space with PSTM_4, the temporal evolution of both the advantages and limitations of machine learning in hydrological modeling becomes clear. Notably, discourse on both these themes has increased significantly since 2021, which reflects the accelerating adoption of AI techniques in hydrological modeling. Analyzing this year-over-year variation allows researchers to gain valuable insights into research focuses and emerging uncertainties across different fields.

Moreover, semantic search offers a valuable approach for academic image retrieval. Utilizing the association between images and their corresponding captions, user prompts can be matched against a large corpus of captions to identify relevant images. I have also compiled a substantial dataset of high-resolution images (about 2 million) extracted from papers listed in Tab.1. Illustrative examples of software applications for image retrieval are provided in Figures S10–S11. This approach holds potential benefits for students, educators, and researchers.

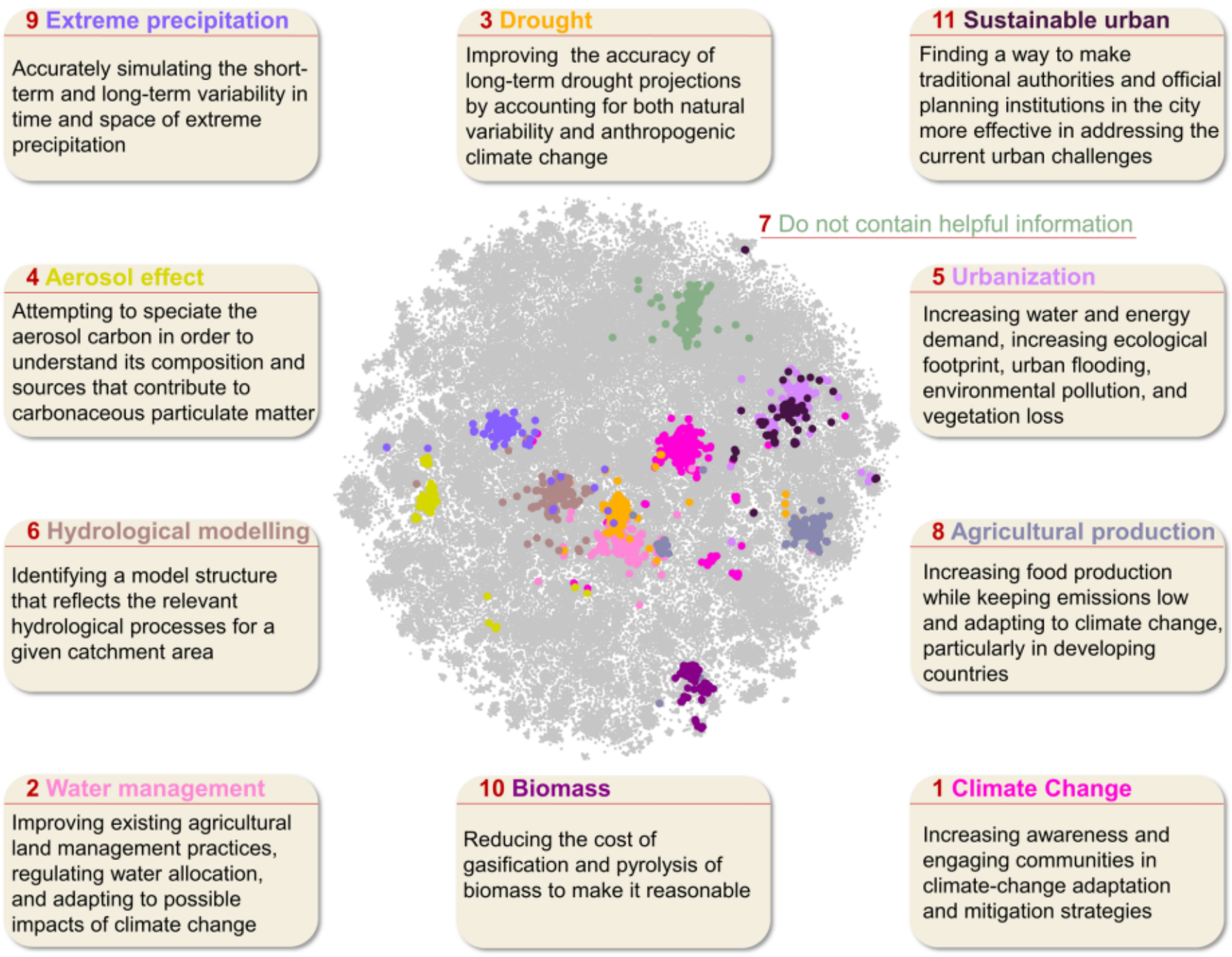

**Figure 4.** The application of sentence-level unsupervised clustering is illustrated through an analysis of texts containing the word "challenge" (case-insensitive). Unsupervised clustering was applied to 271,214 related records, resulting in the selection of the top 11 clusters (visualized as colored points in the figure). Sentences were embedded using PSTM_1, and the minimum similarity score is 0.7 for clustering. A pre-trained LLM (Llama 2 7B) was then used to summarize each cluster's content using the prompt, "Please summarize the challenge." The 11 identified topics and their corresponding challenges are displayed around the scatter plot. Notably, cluster 7 predominantly contains short, declarative sentences with limited information.

### 3.2 Sentence-level unsupervised clustering

As depicted in Figure 4, I filtered a corpus comprising approximately 271,000 sentences to discern researcher perspectives regarding both historical and contemporary challenges. The ten most significant challenges identified by unsupervised clustering primarily revolve around concerns related to the risks of climate change and problems concerning water resources. As underscored by UN-Water (2021), the global community is not on track to achieve critical Sustainable Development Goals (SDGs), especially SDG 6. Worryingly, nearly 80% of the global population is facing significant threats to water security, with one in three individuals lacking access to safe drinking water (Vörösmarty *et al.*, 2010). These issues are further intensified by the consequences of climate change (Ren *et al.*, 2024).

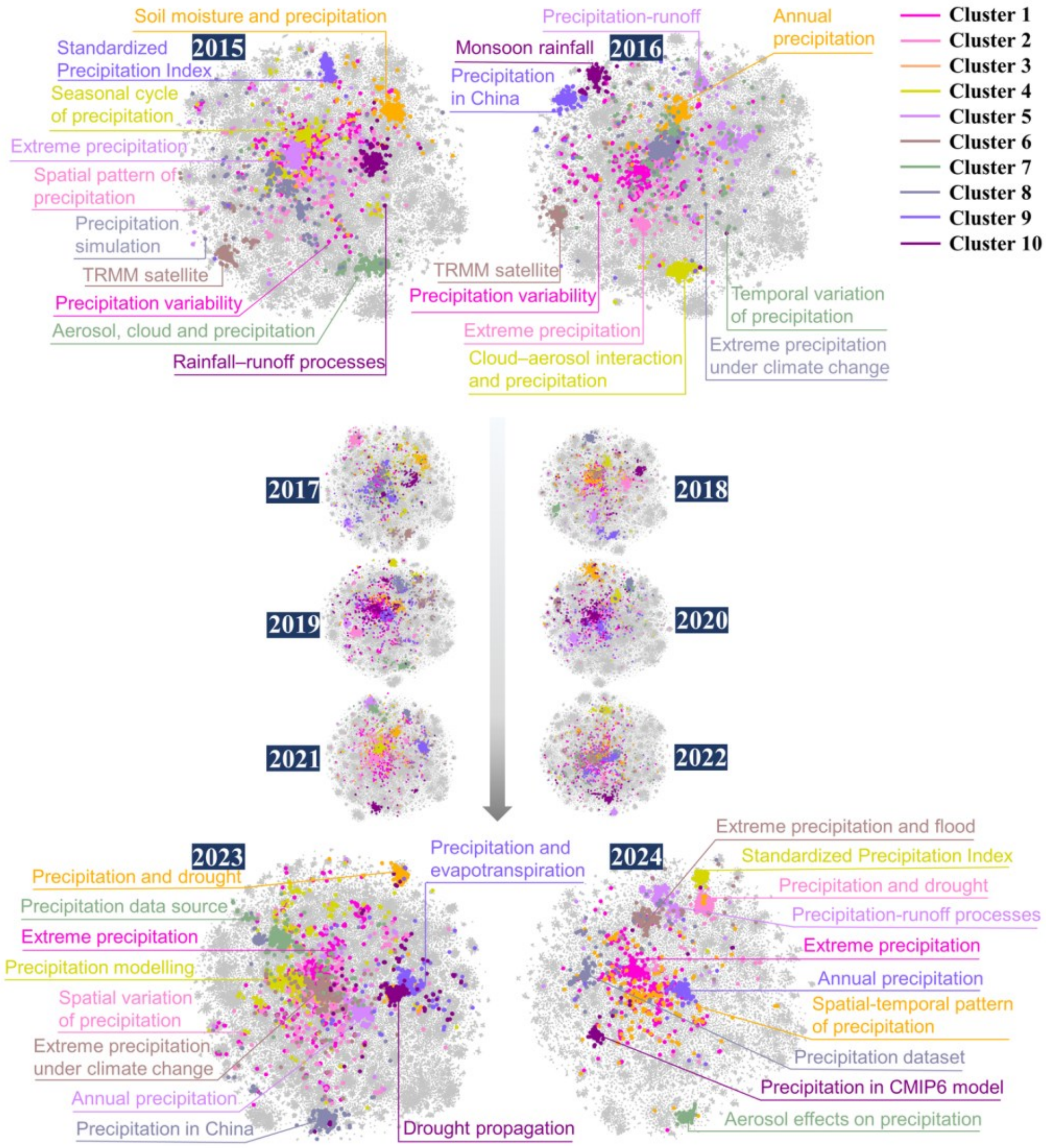

**Figure 5.** Taking "precipitation" as an example to illustrate sentence-level clustering. In this instance, sentences containing the words "precipitation" (case-insensitive) or " rain" (case-insensitive) were extracted, followed by the application of unsupervised clustering for each year. The sub-plot illustrates the dynamic research hotspots within the domain of precipitation from 2015 to 2024. Annually, approximately 130,000 scattered data points are recorded from 2015 to 2023, with this number decreasing to around 50,000 points in 2024 (up to June 2024). The analysis identified the top ten clusters, with "Cluster 1" encompassing the highest number of sentences, gradually decreasing to "Cluster 10". The principal topics within each cluster are presented adjacent to the clusters, maintaining consistency through the use of a uniform color scheme. Due to layout constraints, the principal topics for years from 2017 to 2022 were not displayed.

Figure 5 presents the results of a clustering analysis applied to over 1.6 million precipitation-related sentences, extracted and analyzed on an annual scale. In 2015, the primary emphasis of research was on the temporal and spatial patterns of precipitation and the mechanisms of its formation. By contrast, by 2023 and 2024, the research focus had evolved toward extreme hydrological events (like extreme precipitation, droughts, and floods), the role of precipitation within the hydrosphere, and the effects of climate change on precipitation. This indicates that unsupervised clustering can effectively and automatically identify research hotspots and their trends over time.

### 3.3 Sentence-level sentiment analysis

The critical nature of the ongoing water crisis within the Earth-human system is clearly depicted in Figure 4. I further filtered all sentences containing “water resource” and subjected them to sentiment analysis using *go-emotion* language model. Following the removal of “neutral” sentiment, “approval” was the most frequently observed emotion, particularly in the context of discussions on groundwater management, extreme hydrological events, wastewater treatment, snow-glacier dynamics, and multidisciplinary investigations into water resource management (Figure 6). The analysis further demonstrated that “disappointment,” the second most common emotion, was prominent within conversations pertaining to agricultural drought, groundwater over-extraction, the increasing impact of population and economic expansion on water demand, water pollution, and the intensifying influence of climate change on water resources. While more than 500 records exhibited “optimism,” deeper examination revealed that this sentiment was frequently associated with two central topics: authors highlighting the significance of their studies and admissions that climate change could potentially worsen water-related problems (which would be more appropriately categorized as pessimism). The continued deterioration of water resources despite widespread research indicates a separation between research findings and policy implementation. In general, the overarching sentiment pertaining to water resources is negative, underscoring several critical challenges that necessitate prompt action.

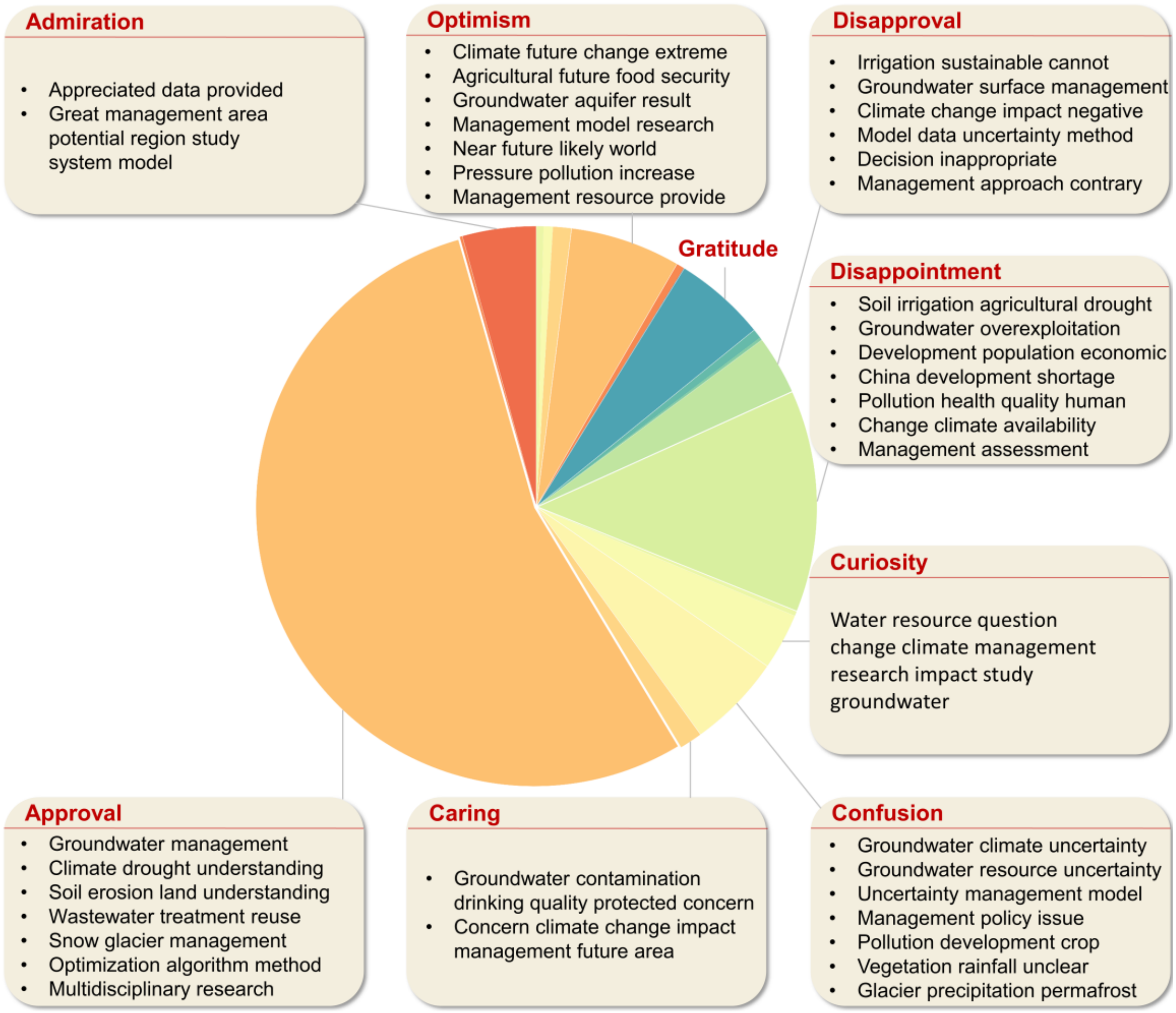

**Figure 6.** Illustration the application of sentence-level sentiment analysis by the *go-emotion* language model, using “water resource” (case-insensitive) as a case study. A total of 98,681 relevant sentences were extracted. Following this, emotional categories with more than 100 supporting sentences were filtered, while excluding neutral emotions. This procedure identified “approval” (4,787) and “disappointment” (1,133) as the most prevalent emotions. The “gratitude” category, primarily consisting of funding acknowledgements, was removed from further analyses. In the topic modeling by the *BERTopic* language model, the “curiosity” sentiment was consolidated into a single topic group. To maintain brevity, only the seven most prominent topic groups identified within the “approval,” “disappointment,” “optimism,” and “confusion” emotional categories are presented.

Furthermore, unsupervised clustering of all negatively-toned freshwater-related sentences (164,381 records in total) detected by the *Twitter-RoBERTa-base-sentiment-latest* language model yielded 40 distinct classes since 2015, coinciding with the introduction of SDGs (Figure 7a; Table S1). These clusters reveal that freshwater ecosystems are under severe stress from interconnected issues, including pollution, climate change, invasive species, overexploitation, and habitat loss, ultimately leading to biodiversity decline and ecosystem degradation. Freshwater scarcity is a worsening global problem due to rising demand, overuse, poor management, industrialization, erratic rainfall patterns, and climate change (e.g., Lazin *et al*., 2020; Sohi *et al*., 2024). This problem increasing in Africa, Asia, the Mediterranean, and on small islands (IPCC, 2022). Alarmingly, projections indicate that 75% of the world's population may face water scarcity by 2050 (Zhang *et al*., 2022). Moreover, microplastics and diverse pollutants from industrial, agricultural, and urban activities contaminate freshwater globally, threatening water quality, aquatic life, and human health (e.g., Chen *et al*., 2022).

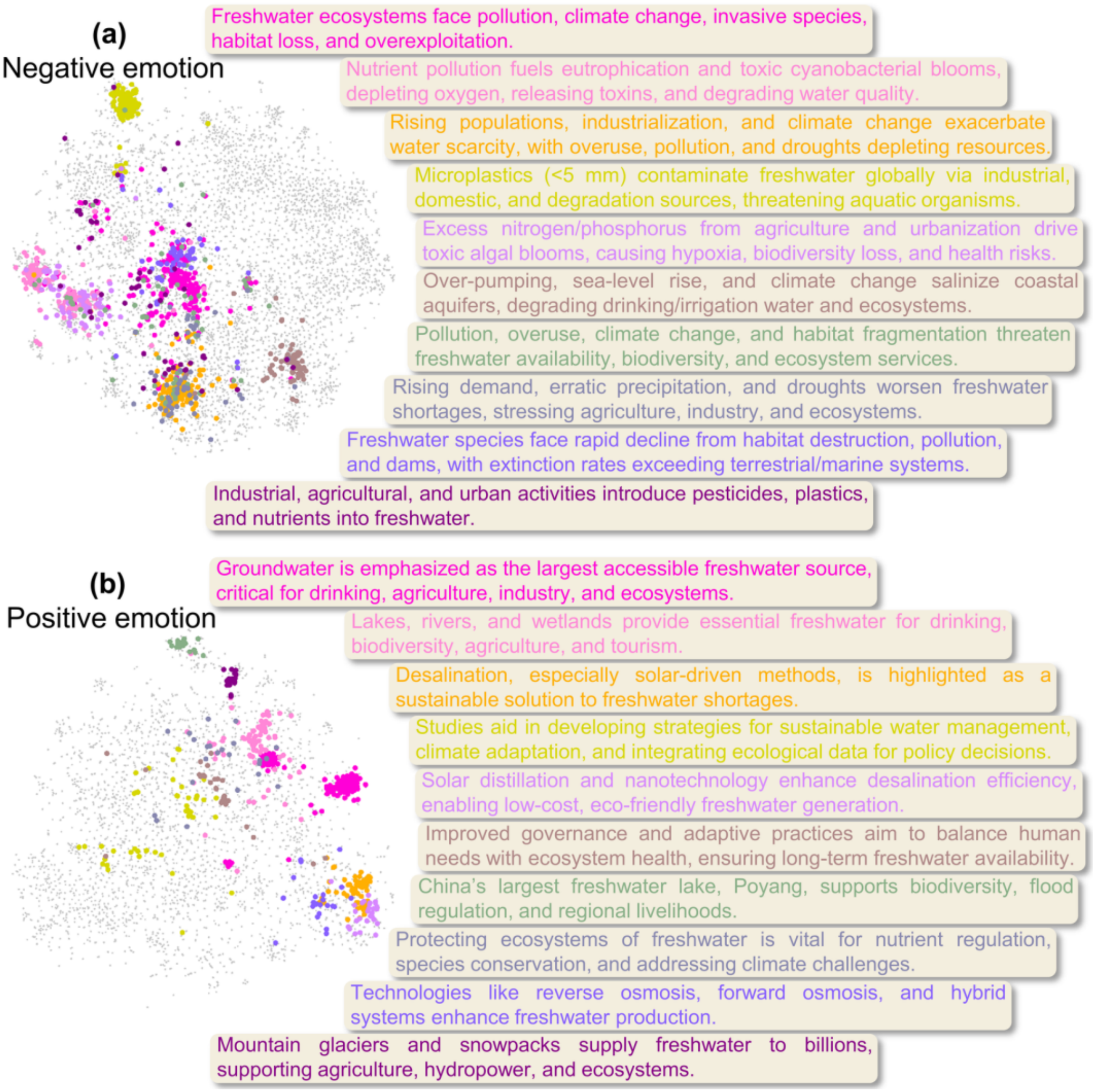

**Figure 7.** Top ten negative (a) and positive (b) emotions, which are detected by the *Twitter-RoBERTa-base-sentiment-latest* language model, in freshwater-related corpus since 2015. (a) is based on Table S1 and (b) is based on Table S2. The dispersed colored points in the scatter plot correspond directly to the colored LLM summarization on the right.

In contrast, analysis of the positively-toned freshwater corpus (Fig. 7b; Table S2) indicates that the importance of groundwater, lakes, rivers, wetlands, mountain glaciers, and snowpacks as sources of freshwater is frequently emphasized. Desalination technologies, particularly the solar distillation, the reverse osmosis, and the multi-effect distillation technologies, offer sustainable approaches to freshwater production (e.g., Forootan & Forootan, 2024). Furthermore, the efficiency of solar desalination system can be improved by energy storage materials including nanotechnology and phase change materials (Jamil *et al*., 2023). To achieve SDGs, sustainable freshwater management is essential.

## 4 Conclusion and discussion

Addressing the rapid expansion of literature, the need for effective interdisciplinary and transdisciplinary knowledge transfer, and the computation cost and the black-box nature posed by LLMs are critical for science community.

In this perspective, I establish a framework for a precise and cost-effective information acquisition in geoscience by integrating a range of MiniLMs with an extensive and meticulously curated corpus of high-quality sentences sourced from well-respected journals, thereby substantially enhancing the applicability of MiniLMs within geoscience. Through the creation of a corpus of approximately 77 million sentences, my work effectively demonstrates the capability of MiniLMs across three key downstream tasks: sentence-level semantic search, sentence-level unsupervised clustering, and sentence-level sentiment analysis.

Specifically, the semantic search facilitates the fact retrieval of a particular subdomain, efficiently converting a simple query into a detailed, professional, and often extensive summary when coupled with high-performance LLMs. Furthermore, this methodology aids in elucidating the inherent complexity, contradictoriness, and temporal evolution of given subjects across numerous disciplines. Importantly, it offers a cost-effective means of locating domain-specific knowledge with quantified data, potentially exceeding the reliability of powerful commercial tools like ChatGPT-4. In summary, precise and cost-effective information retrieval from geoscience literature can be achieved by: (1) curating a comprehensive collection of well-respected publications within a specific field; (2) segmenting these publications into high-quality individual sentences; and (3) utilizing freely available MiniLMs, such as *all-MiniLM-L6-v2*.

Unsupervised clustering can automatically identify similar patterns in sentences, aiding in the discovery of research priorities, conclusions, and limitations within a field or a discipline. This enables the formation of an overall disciplinary perspective, independent of subjective human judgment, by automatically capturing structured knowledge. However, unstructured knowledge, which is not captured by clustering, could be equally valuable in my opinion. These non-mainstream or marginalized viewpoints merit further re-evaluation and could ultimately drive the progress of the discipline.

Current sentiment analysis models, basically trained on social media data, this perspective remains exploratory in its application to the analysis of specific scientific issue. However, with a sufficiently large corpus, negative emotions detected by sentiment analysis can comprehensively expose the challenges, causes, and consequences associated with specific issues. This enables a thorough examination of critical issues facing the Earth-human system, such as climate change, water resource shortage, ecosystem degradation, pollution, agricultural security, and rapid urbanization and industrialization. As researchers, our more crucial task might be to seek potential solutions within crises. While positive emotions revealed by sentiment analysis can provide inspiration, they are unlikely to fully resolve these challenges.

Despite its strengths, the extraction of a high-quality corpus from PDFs is still subject to several limitations. The corpus is affected by fragmented sentences and words, exemplified by the misinterpretation of “significant” as “signifi cant” due to typesetting inconsistencies. Furthermore, the correct handling of formulas remains a substantial challenge. Additionally, the necessity to uphold the copyright of the literature restricts this methodology to research-oriented uses within the science community. Also, limited by my background, the examples provided in this perspective are primarily focused on atmospheric sciences.

## 5 Using MiniLMs to validate LLMs

The current focus within the research community is largely directed toward LLMs. The escalating capabilities of LLMs, while largely facilitating knowledge exchange, might also significantly erode public confidence in knowledge creation. If LLMs were to dominate the interpretation of discourse within human civilization, what would remain of the unique value of humanity? In contrast, MiniLM offers individuals the capacity to filter and reorganize information, and even question the outputs of LLMs, offering high precise at a much lower computational cost. Moreover, leveraging the ancient Chinese wisdom of Yin-Yang (阴-阳) philosophy, such as through completely opposite prompts or sentiment analysis, it is relatively helpful to discover and understand the duality of things, and fully recognize the progress and limitations of the discipline.

Furthermore, a user-friendly sentence-level MiniLM-based search platforms provided by reputable publishers or groups would be invaluable, allowing different groups, regardless of their expertise level, to access relevant professional information. The increasing prevalence of open-access publishing suggests that authors might prioritize a broad attention. This raises another fundamental question: Is the value of academic papers measured by citation counts, or should they be considered a collective body of knowledge of human serving the common good? For Earth science researchers, what constitutes "scientific importance"? Is it primarily the health of our planet or the well-being of humanity? Despite the extensive body of research emerging from Earth science, the unprecedented rates of change in Earth system is still likely exceeding "safe" levels in several dimensions (Anenberg *et al*., 2019; Rockström *et al*., 2021). A more holistic approach is needed, one that transcends disciplinary boundaries and incorporates multi-dimensional perspectives. Research papers, representing the collective wisdom of Earth, environment, and space scientists, should be readily accessible to the public, policymakers, and even middle school students.

## Acknowledgements

The author would like to acknowledge the National Natural Science Foundation of China under Grant 42205074. I am deeply grateful to Pro. Michael Wysession for providing the opportunity to submit this work for review. I also extend my sincere appreciation to Pro. Yunyue Elita Li and the two anonymous reviewers for their insightful and constructive feedback.

## Data Availability Statement

A simplified procedure for utilizing MiniLM is available at Zhang (2025).

## Conflict of Interest

The author declares no conflicts of interest relevant to this study

## References

Anenberg, S. C., Dutton, A., Goulet, C., Swain, D. L., & van der Pluijm, B. (2019). Toward a resilient global society: Air, sea level, earthquakes, and weather. *Earth's Future*, *7*, 854–864. https://doi.org/10.1029/2019EF001242

Bergen, K. J., Johnson, P. A., de Hoop, M. V. & Beroza, G. C. (2019). Machine learning for data-driven discovery in solid Earth geoscience. *Science*, *363*, eaau0323. https://doi.org/10.1126/science.aau0323

Best, J. (2019). Anthropogenic stresses on the world's big rivers. *Nature Geoscience*, *12*(1), 7–21. https://doi.org/10.1038/s41561-018-0262-x

Chen, H., Jia, Q., Sun, X., Zhou, X., Zhu, Y., Guo, Y., & Ye, J. (2022). Quantifying microplastic stocks and flows in the urban agglomeration based on the mass balance model and source-pathway-receptor framework: Revealing the role of pollution sources, weather patterns, and environmental management practices. *Water Research*, 224, 119045. https://doi.org/10.1016/j.watres.2022.119045

Cui, H. (2021). Inside out: deep carbon linked to deep-time carbon cycle. *Science Bulletin*, *66*(18), 1822–1824. https://doi.org/10.1016/j.scib.2021.06.001

Demszky, D., Movshovitz-Attias, D., Ko, J., Cowen, A., Nemade, G., & Ravi, S. (2020). GoEmotions: A dataset of fine-grained emotions. arXiv preprint arXiv:2005.00547.

Forootan, M. M., & Ahmadi, A. (2024). Machine learning-based optimization and 4E analysis of renewable-based polygeneration system by integration of GT-SRC-ORC-SOFC-PEME-MED-RO using multi-objective grey wolf optimization algorithm and neural networks. *Renewable and Sustainable Energy Reviews*, *200*, 114616. https://doi.org/10.1016/j.rser.2024.114616

Foroumandi, E., Moradkhani, H., Sanchez-Vila, X., Singha, K., Castelletti, A., & Destouni, G. (2023). ChatGPT in hydrology and Earth sciences: Opportunities, prospects, and concerns. *Water Resources Research*, *59*, e2023WR036288. https://doi.org/10.1029/2023WR036288

Gougherty, A. V., & Clipp, H. L. (2024). Testing the Reliability of an AI-Based Large Language Model to Extract Ecological Information From the Scientific Literature. *NPJ Biodiversity*, *3*, 1–5. https://doi.org/10.1038/s44185-024-00043-9

IPCC. (2022). In H. O. P.rtner, D. C. Roberts, M. Tignor, E. S. Poloczanska, K. Mintenbeck, A. Alegría, et al. (Eds.), Climate change 2022:Impacts, adaptation, and vulnerability. Cambridge University Press. https://dx.doi.org/10.1017/9781009325844

Jablonka, K. M., Schwaller, P., Ortega-Guerrero, A., & Smit, B. (2024). Leveraging large language models for predictive chemistry. *Nature Machine Intelligence*, *6*, 161–169. https://doi.org/10.1038/s42256-023-00788-1

Jamil, F., Hassan, F., Shoeibi, S., & Khiadani, M. (2023). Application of advanced energy storage materials in direct solar desalination: a state of art review. *Renewable and Sustainable Energy Reviews*, *186*, 113663. https://doi.org/10.1016/j.rser.2023.113663

Lazin, R., Shen, X., Koukoula, M., & Anagnostou, E. (2020). Evaluation of the hyper-resolution model-derived water cycle components over the upper Blue Nile Basin. *Journal of Hydrology*, *590*, 125231. https://doi.org/10.1016/j.jhydrol.2020.125231

Li, Y. E., O'Malley, D., Beroza, G., Curtis, A., & Johnson, P. (2023). Machine learning developments and applications in Solid-Earth geosciences: Fad or future? *Journal of Geophysical Research: Solid Earth*, *128*, e2022JB026310. https://doi.org/10.1029/2022JB026310

Kononova, O., He, T., Huo, H., Trewartha, A., Olivetti, E. A., & Ceder, G. (2021). Opportunities and challenges of text mining in materials research. *Iscience*, *24*(3). https://doi.org/10.1016/j.isci.2021.102155

M. Bran, A., Cox, S., Schilter, O., Baldassari, C., White, A. D., & Schwaller, P. (2024). Augmenting large language models with chemistry tools. *Nature Machine Intelligence*, *6*(5), 525–535. https://doi.org/10.1038/s42256-024-00832-8

Meng, R., Liu, Y., Joty, S. R., Xiong, C., Zhou, Y., & Yavuz, S. (2024). Sfrembedding-mistral: enhance text retrieval with transfer learning. Salesforce AI Research Blog, 3, 6.

Nielsen, J. Ø., & D'haen, S. A. L. (2014). Asking about climate change: Reflections on methodology in qualitative climate change research published in global environmental change since 2000. *Global Environmental Change*, *24*, 402–409. https://doi.org/10.1016/j.gloenvcha.2013.10.006

Ott, S., Hebenstreit, K., Liévin, V., Hother, C. E., Moradi, M., Mayrhauser, M. , Robert, P., Ole, W., & Samwald, M. (2023). ThoughtSource: A central hub for large language model reasoning data. *Scientific Data*, *10*, 528. https://doi.org/10.1038/s41597-023-02433-3

Ramachandran, R., & Bugbee, K. (2025). Balancing practical uses and ethical concerns:The role of large language models in scientific research. *Perspectives of Earth and Space Scientists*, *6*, e2024CN000258. https://doi.org/10.1029/2024CN000258

Reimers, N., & Gurevych, I. (2019a). Sentence-bert: Sentence embeddings using siamese bert-networks. arXiv preprint arXiv:1908.10084.

Reimers, N., & Gurevych, I. (2019b). Sentence Transformers documentation. Accessed at 21 July 2025. https://www.sbert.net/docs/sentence_transformer/pretrained_models.html

Ren, Y., Yu, H., Huang, J., Peng, M., & Zhou, J. (2024). The projected response of the water cycle to global warming over drylands in East Asia. *Earth's Future*, *12*, e2023EF004008. https://doi.org/10.1029/2023EF004008

Rockström, J., Gupta, J., Lenton, T. M., Qin, D., Lade, S. J., Abrams, J. F., et al. (2021). Identifying a safe and just corridor for people and the planet. *Earth's Future*, *9*, e2020EF001866.https://doi.org/10.1029/2020EF001866

Sohi, H. Y., Zahraie, B., Dolatabadi, N., & Zebarjadian, F. (2024). Application of VIC-WUR model for assessing the spatiotemporal distribution of water availability in anthropogenically-impacted basins. *Journal of Hydrology*, *637*, 131365. https://doi.org/10.1016/j.jhydrol.2024.131365

Tshitoyan, V., Dagdelen, J., Weston, L., Dunn, A., Rong, Z., Kononova, O., Persson, K. A., Ceder, G., & Jain, A. (2019). Unsupervised word embeddings capture latent knowledge from materials science literature. *Nature*, *571*(7763), 95–98. https://doi.org/10.1038/s41586-019-1335-8

UN-Water. (2021). Summary Progress Update 2021: SDG 6 — Water and sanitation for all, Geneva. https://www.unwater.org/publications/sdg-6-progress-reports

Vance, T. C., Huang, T., & Butler, K. A. (2024). Big data in Earth science: Emerging practice and promise. *Science*, *383*(6688), eadh9607. https://doi.org/10.1126/science.adh9607

Vörösmarty, C. J., McIntyre, P. B., Gessner, M. O., Dudgeon, D., Prusevich, A., Green, P., Glidden, S., Bunn, S. E., Sullivan, C. A., Reidy Liermann, C., & Davies, P. (2010). Global threats to human water security and river biodiversity. *Nature*, *467*(7315), 555–561. https://doi.org/10.1038/nature09440

Wu, H., Yang, Q., Liu, J., & Wang, G. (2020). A spatiotemporal deep fusion model for merging satellite and gauge precipitation in China. *Journal of Hydrology*, *584*, 124664. https://doi.org/10.1016/j.jhydrol.2020.124664

Zhang, J. (2025). Tips for MiniLM [Dataset]. Retrieved from https://doi.org/10.6084/m9.figshare.29616506.v1

Zhang, W., Huang, J., Zhang, T., & Tan, Q. (2022). A risk-based stochastic model for supporting resources allocation of agricultural water-energy-food system under uncertainty. *Journal of Hydrology*, *610*, 127864. https://doi.org/10.1016/j.jhydrol.2022.127864

## Supporting Information

**Table S1.** *Unsupervised clustering was performed on negative emotions in freshwater-related corpus since 2015. Sentences were embedded using PSTM_4, with clustering parameters set to a minimum cluster count of 10 and a minimum similarity score of 0.85. Cluster topics and summaries were automatically generated by the DeepSeek LLM, using the prompt: "Please find the topic within 10 words and summarize the text file within 50 words."*

| Cluster | Topic | Summary |
|---|---|---|
| 0 | **Threats to freshwater ecosystems: pollution, climate change, habitat loss** | Freshwater ecosystems face pollution, climate change, invasive species, habitat loss, and overexploitation, driving biodiversity decline, hypoxia, and ecosystem service degradation. Urgent action is needed to mitigate these interconnected stressors threatening aquatic life and human well-being. |
| 1 | **Eutrophication and harmful algal blooms threatening freshwater ecosystems** | Nutrient pollution fuels eutrophication and toxic cyanobacterial blooms, depleting oxygen, releasing toxins, and degrading water quality. These blooms harm biodiversity, disrupt food chains, and jeopardize drinking water, fisheries, and human health globally. |
| 2 | **Freshwater scarcity driven by population growth and climate change** | Rising populations, industrialization, and climate change exacerbate water scarcity, with overuse, pollution, and droughts depleting resources. By 2025, half the global population may face water stress, threatening agriculture, health, and socioeconomic stability. |
| 3 | **Microplastic pollution in freshwater ecosystems** | Microplastics (<5 mm) contaminate freshwater globally via industrial, domestic, and degradation sources, threatening aquatic organisms through ingestion, toxin adsorption, and ecosystem disruption, with risks to biodiversity and human health via food chains. |
| 4 | **Nutrient pollution causing harmful algal blooms** | Excess nitrogen/phosphorus from agriculture and urbanization drive toxic algal blooms in freshwater, causing hypoxia, biodiversity loss, and health risks via contaminated water and seafood. |
| 5 | **Seawater intrusion threatening coastal freshwater resources** | Over-pumping, sea-level rise, and climate change salinize coastal aquifers, degrading drinking/irrigation water and ecosystems, exacerbating scarcity in densely populated regions. |
| 6 | **Multiple stressors degrading freshwater sustainability** | Pollution, overuse, climate change, and habitat fragmentation threaten freshwater availability, biodiversity, and ecosystem services, necessitating urgent integrated management. |
| 7 | **Climate change and population growth intensifying water scarcity** | Rising demand, erratic precipitation, and droughts worsen freshwater shortages, stressing agriculture, industry, and ecosystems, particularly in arid regions. |
| 8 | **Biodiversity loss in freshwater ecosystems** | Freshwater species face rapid decline from habitat destruction, pollution, and dams, with extinction rates exceeding terrestrial/marine systems, disrupting ecological functions. |
| 9 | **Anthropogenic pollutants degrading freshwater quality** | Industrial, agricultural, and urban activities introduce pesticides, plastics, and nutrients into freshwater, compromising ecosystems, human health, and water security. |
| 10 | **Toxic cyanobacterial blooms in freshwater systems** | Nutrient-driven cyanobacteria blooms release toxins (e.g., microcystins), endangering drinking water, aquatic life, and human health through oxygen depletion and poisoning. |
| 11 | **Global freshwater scarcity crisis** | Population growth, industrialization, and pollution drive global freshwater shortages, affecting billions. Projections indicate 75% of people may face scarcity by 2050, threatening food, energy security, and ecosystems. |
| 12 | **Human impacts on freshwater biodiversity** | Urbanization, agriculture, and pollution degrade freshwater ecosystems, causing biodiversity loss and population declines in invertebrates and fish through habitat alteration and contamination. |
| 13 | **Freshwater salinization risks** | Rising salinity from road salts, wastewater, and agriculture harms aquatic life, mobilizes contaminants, and threatens drinking water, biodiversity, and ecosystem services globally. |
| 14 | **Climate change threats to freshwater** | Warming temperatures, altered flow regimes, and stratification degrade freshwater ecosystems, stressing fish populations and increasing risks of hypoxia and harmful algal blooms. |
| 15 | **Coastal aquifer seawater intrusion** | Over-pumping and sea-level rise cause saltwater intrusion into coastal aquifers, contaminating freshwater supplies, harming agriculture, and endangering drinking water quality. |
| 16 | **Emerging contaminants in freshwater** | Microplastics, pharmaceuticals, and pollutants accumulate in freshwater ecosystems, with poorly understood ecological and health risks due to limited research and monitoring. |
| 17 | **Depletion of freshwater resources** | Urbanization, climate change, and rising demand deplete freshwater availability, exacerbating water stress and threatening sustainability in arid and densely populated regions. |
| 18 | **Nutrient-driven freshwater eutrophication** | Excess phosphorus and nitrogen from agriculture and wastewater fuel harmful algal blooms, degrading water quality, causing hypoxia, and disrupting aquatic ecosystems globally. |
| 19 | **Irrigation strains on freshwater** | Agricultural expansion intensifies freshwater extraction, depleting aquifers and rivers, worsening scarcity, and heightening competition in water-stressed regions. |
| 20 | **Climate-driven saltwater intrusion** | Sea-level rise, droughts, and over-pumping accelerate saltwater intrusion into coastal freshwater, contaminating supplies and disrupting ecosystems and human |
| | | livelihoods. |
| 21 | **Industrial pollution and heavy metals degrading freshwater resources** | Industrial activities and heavy metals (e.g., $Hg^{2+}$) contaminate freshwater, causing scarcity and threatening ecosystems, particularly in developing nations. |
| 22 | **Freshwater biodiversity loss exceeds marine/terrestrial ecosystems** | Freshwater species face rapid declines due to habitat loss, pollution, and invasive species, with extinction rates surpassing marine/terrestrial environments. |
| 23 | **Agricultural runoff and nitrogen pollution harm freshwater quality** | Nitrogen from agriculture causes eutrophication, oxygen depletion, and ecosystem damage in freshwater, impacting drinking water and coastal areas. |
| 24 | **Freshwater scarcity due to supply-demand mismatches and overuse** | Over-extraction, droughts, and poor infrastructure exacerbate freshwater shortages, stressing regions like the YRB and arid zones. |
| 25 | **Sea-level rise contaminates freshwater via saltwater intrusion** | Coastal aquifers face salinization from rising seas, threatening freshwater availability for ecosystems, agriculture, and human consumption. |
| 26 | **Pesticide pollution disrupts freshwater biodiversity** | Agricultural pesticides harm zooplankton and fish, reducing biodiversity and causing ecological risks, particularly in vulnerable regions. |
| 27 | **Climate change intensifies freshwater system instability** | Extreme weather and pollution worsen freshwater scarcity, eutrophication, and ecosystem health, challenging global water security. |
| 28 | **Excess phosphorus drives freshwater eutrophication globally** | Phosphorus from agriculture and industry triggers algal blooms, degrading water quality and causing economic losses in freshwater systems. |
| 29 | **Toxic contaminants threaten freshwater ecosystems and human health** | MPs, heavy metals, and pharmaceuticals bioaccumulate, harming aquatic life and human health through toxicological effects. |
| 30 | **Microplastics severely impact freshwater ecosystems** | MPs disrupt organisms, nutrient cycles, and food webs in freshwater, with risks from additives and cross-system contamination. |
| 31 | **Agriculture's high freshwater consumption threatens sustainability** | Agriculture uses 50-80% of global freshwater, causing scarcity, ecosystem stress, and competition with municipal/environmental needs. Irrigation-driven depletion, pollution, and climate change exacerbate risks to sustainable food production and water security. |
| 32 | **Data gaps hinder freshwater dynamics understanding** | Limited in-situ data, sparse monitoring, and poor spatial-temporal resolution obscure freshwater storage quantification, seasonal variability, and ecological impacts. Remote sensing and modeling face uncertainties, hampering policy and management decisions. |
| 33 | **Freshwater influx weakens Atlantic Ocean circulation (AMOC)** | Increased freshwater from ice melt reduces North Atlantic Deep Water formation, weakening Atlantic Ocean circulation. This disrupts heat transport, causing regional cooling (e.g., “warming hole”) and amplifying climate feedbacks. |
| 34 | **Plastic pollution endangers freshwater ecosystems globally** | Plastics infiltrate rivers and lakes, harming biodiversity via entanglement, toxicity, and food chain contamination. Agricultural runoff and poor waste management exacerbate pollution, threatening water quality and human health. |
| 35 | **Freshwater ecotoxicity and eutrophication dominate environmental impacts** | Key impact categories include freshwater eutrophication, ecotoxicity, and human toxicity. Agriculture and industry drive nutrient pollution, chemical contamination, and resource depletion, degrading aquatic ecosystems. |
| 36 | **Multiple stressors threaten freshwater biodiversity and ecosystems** | Pollution, warming, invasive species, and habitat loss synergistically harm freshwater organisms. Mollusks and invertebrates face high extinction risks, disrupting food webs and ecosystem resilience. |
| 37 | **Climate change intensifies droughts and freshwater scarcity** | Rising temperatures and erratic rainfall reduce water availability, exacerbating droughts. Regions like the Mediterranean and small islands face agricultural losses, ecosystem collapse, and water conflicts. |
| 38 | **Eutrophication plagues China’s major freshwater lakes** | Taihu, Chaohu, and Poyang Lakes suffer algal blooms and pollution from agriculture, industry, and urbanization. Nutrient runoff degrades water quality, threatening drinking supplies and biodiversity. |
| 39 | **Small islands face acute freshwater scarcity and salinity** | Limited groundwater, climate vulnerability, and overuse cause water shortages in island communities. Saltwater intrusion and droughts jeopardize ecosystems and human survival, requiring urgent sustainable management. |

**Table S2.** *Similar to Tab. S1 but for the positive emotion in freshwater-related corpus.*

| Cluster | Topic | Summary |
|---|---|---|
| 0 | **Groundwater as the primary global freshwater resource** | Groundwater is emphasized as the largest accessible freshwater source, critical for drinking, agriculture, industry, and ecosystems. It supports billions of people and global food security, with studies citing its irreplaceable role in sustaining human needs and environmental stability. |
| 1 | **Lakes and rivers as vital freshwater ecosystems** | Lakes, rivers, and wetlands provide essential freshwater for drinking, biodiversity, agriculture, and tourism. They offer ecosystem services like flood control, nutrient cycling, and cultural value, supporting millions globally while |
| | | balancing ecological and socioeconomic needs. |
| 2 | **Desalination technologies addressing freshwater scarcity** | Desalination, especially solar-driven methods, is highlighted as a sustainable solution to freshwater shortages. Innovations like reverse osmosis and hydrate-based processes aim to efficiently convert seawater into potable water, particularly for arid regions. |
| 3 | **Research insights into freshwater ecosystem management** | Studies focus on understanding freshwater systems, pollution control, and biodiversity conservation. Findings aid in developing strategies for sustainable water management, climate adaptation, and integrating ecological data for policy decisions. |
| 4 | **Solar-driven desalination advancements for freshwater production** | Solar distillation and nanotechnology enhance desalination efficiency, enabling low-cost, eco-friendly freshwater generation. Innovations like interfacial evaporation and phase-change materials improve productivity, addressing water scarcity in energy-limited regions. |
| 5 | **Sustainable management of freshwater resources** | Strategies emphasize holistic conservation, biodiversity protection, and stakeholder collaboration. Improved governance and adaptive practices aim to balance human needs with ecosystem health, ensuring long-term freshwater availability. |
| 6 | **Poyang Lake's ecological and economic significance** | China's largest freshwater lake, Poyang, supports biodiversity, flood regulation, and regional livelihoods. Studies highlight its role in carbon sequestration, fisheries, and as a Ramsar wetland critical to the Yangtze River Basin. |
| 7 | **Freshwater ecosystems' role in global sustainability** | Freshwater environments sustain biodiversity, carbon cycles, and human well-being. Protecting these ecosystems is vital for nutrient regulation, species conservation, and addressing climate challenges through integrated socio-ecological approaches. |
| 8 | **Sustainable freshwater production through energy-efficient technologies** | Technologies like reverse osmosis, forward osmosis, and hybrid systems enhance freshwater production, improve energy efficiency, reduce $CO_2$ emissions, and decrease fossil fuel reliance, addressing future water demands sustainably while promoting renewable energy integration. |
| 9 | **Glaciers and snowpacks as vital freshwater sources** | Mountain glaciers and snowpacks supply freshwater to billions, supporting agriculture, hydropower, and ecosystems in regions like the Himalayas, Tibetan Plateau, and Andes, crucial for arid areas and downstream populations. |
| 10 | **Methane dynamics in freshwater ecosystems and climate impact** | Freshwater ecosystems significantly contribute to global methane emissions, with temperature and eutrophication influencing $CH_4$ production. Improved models are needed to assess their role in greenhouse gas budgets and climate change. |

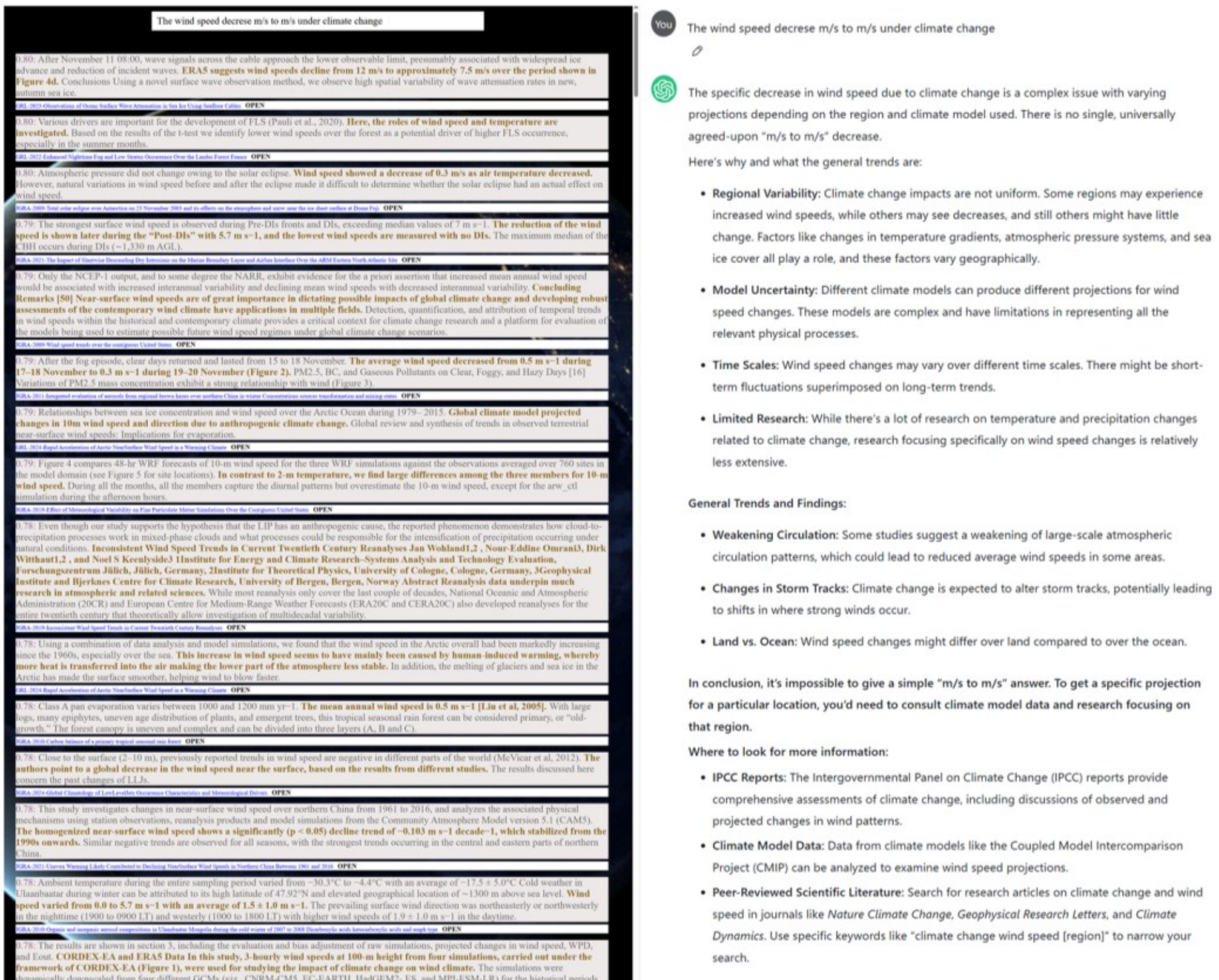

**Figure S1.** A software demonstration illustrates semantic searching capabilities within atmospheric science literature. The software was run on a laptop equipped with an Intel® Core™ i7-10750H CPU @ 2.60GHz, 64GB of memory, and an NVIDIA Quadro P620 GPU. The right panel replicates a query generated by ChatGPT-4: "The wind speed decrease m/s to m/s under climate change."

The top-matched sentences, identified by the PSTM_3 model, are highlighted in orange, with the corresponding relevance score indicated numerically. Contextual sentences preceding and following the matched segments are displayed in black, with blue annotations denoting the original reference. An "OPEN" button facilitates direct access to the original publication in PDF format. In this instance, sentence-based searching yielded results quantifying wind speed decreases. Furthermore, the search highlighted discrepancies in wind speed trends under climate change as reported by different reanalysis products. Due to display constraints, only the highest-ranked sentences are presented. In contrast, ChatGPT-4 suggested consulting publications in journals such as *Geophysical Research Letters* and *Nature Climate Change*, offering a broader but less specific result.

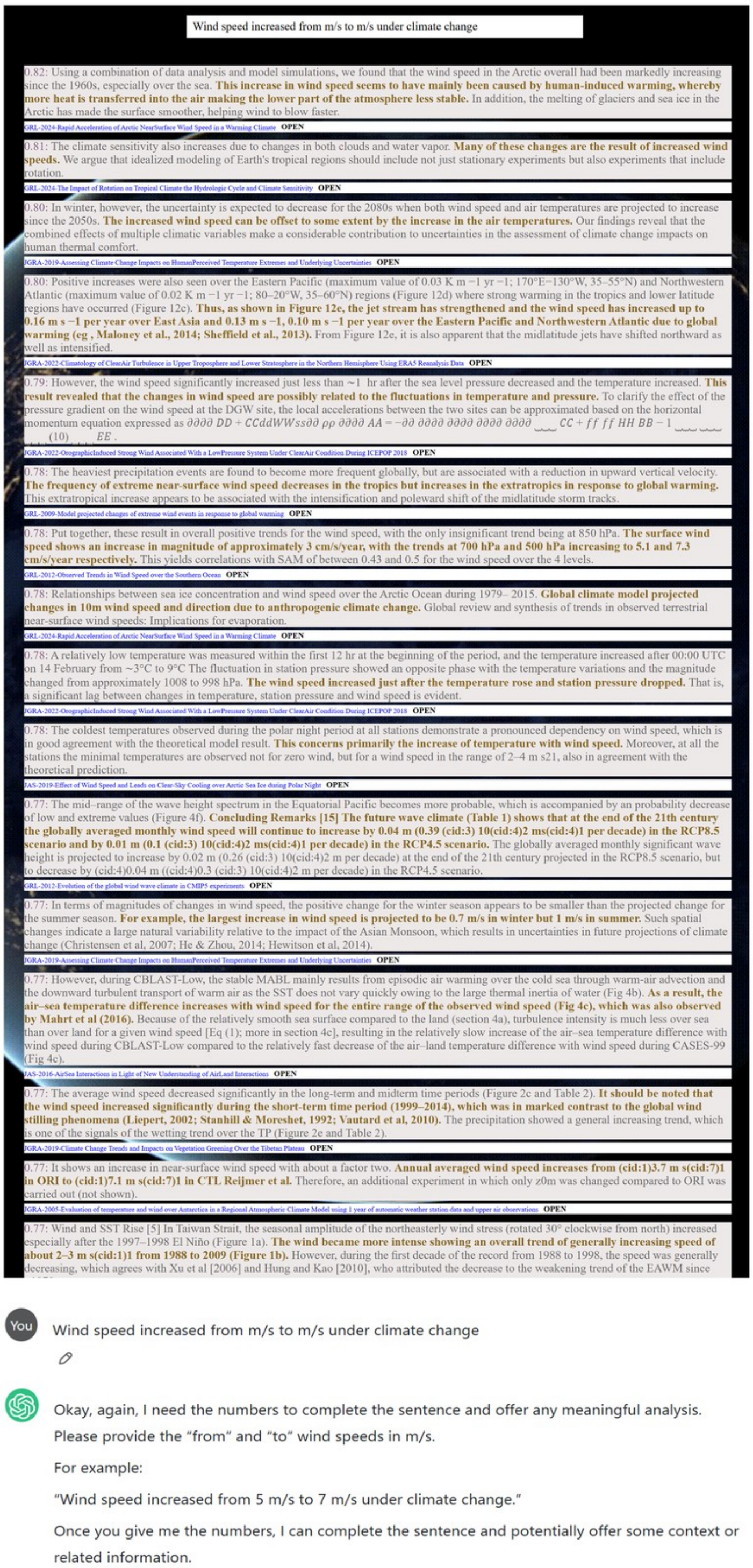

**Figure S2.** Conversely, applying the query "Wind speed increased from m/s to m/s under climate change" to the semantic search software reveals substantial evidence indicating an increase in wind speed (near-surface winds or jet stream) within historical research. This includes quantified data and associated mechanisms. In contrast, ChatGPT-4 was unable to retrieve quantitative data pertaining to wind speed increases.

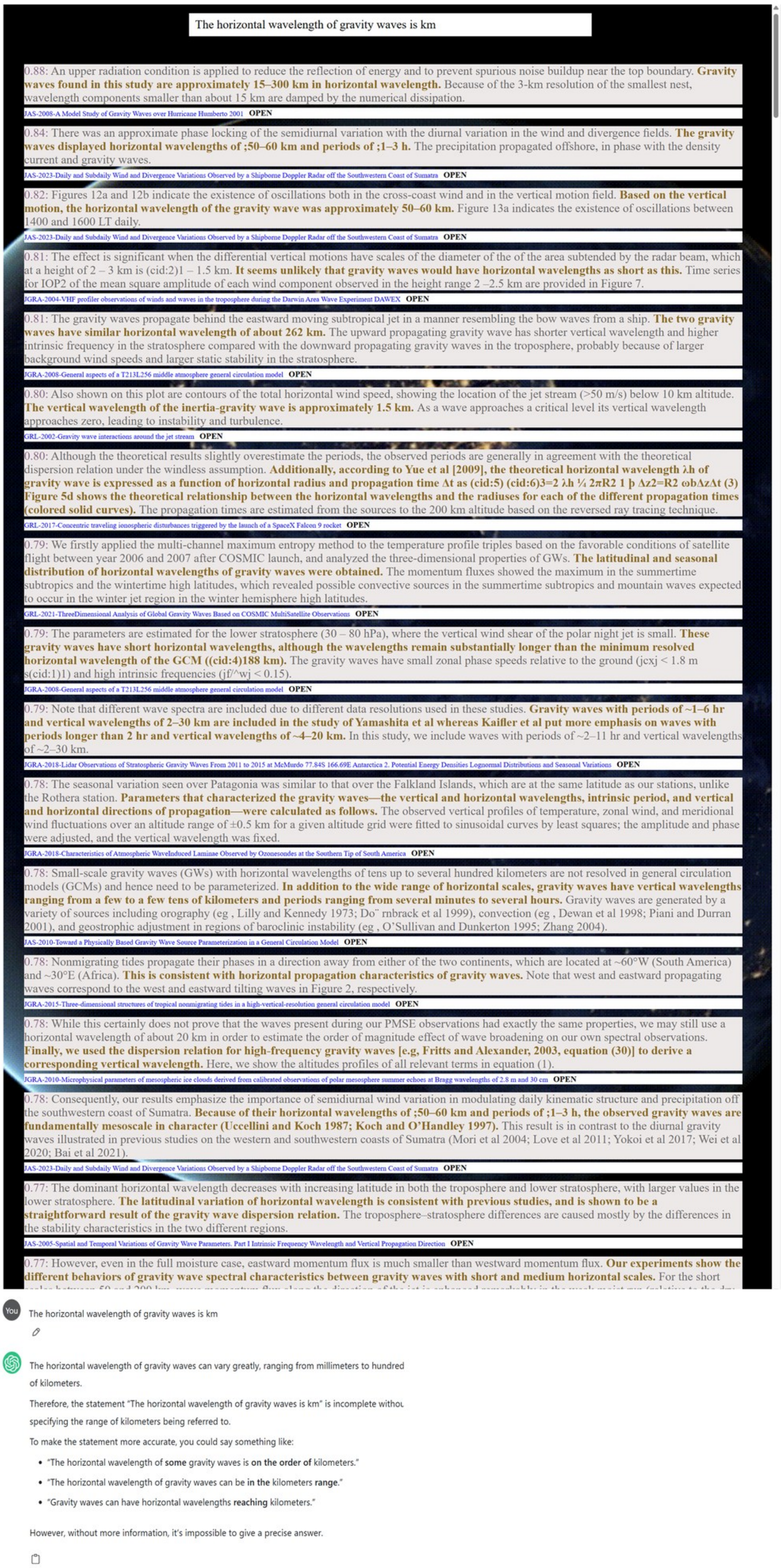

**Figure S3.** A simple query, "The horizontal wavelength of gravity waves is km" was conducted. The software generated a range of gravity wave parameters, including vertical and horizontal wavelengths, as well as wave periods. In contrast, ChatGPT-4 stated that wavelengths range from millimeters to hundreds of kilometers, a conclusion obviously inconsistent with current research.

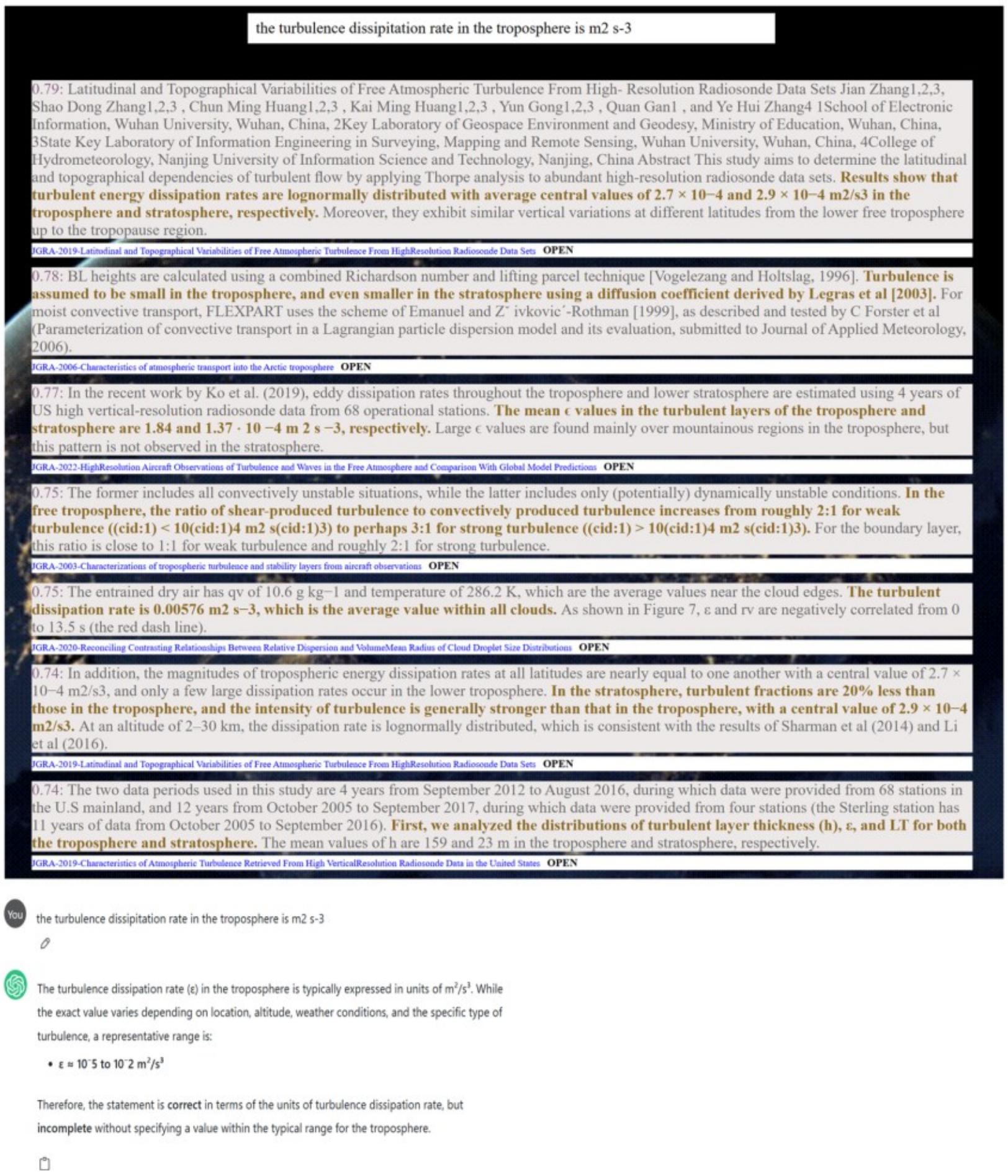

**Figure S4.** The query "the turbulence dissipation rate in the troposphere is m2 s-3" was submitted. The initial output from software quantitatively characterized turbulence intensity in both the troposphere and the stratosphere.

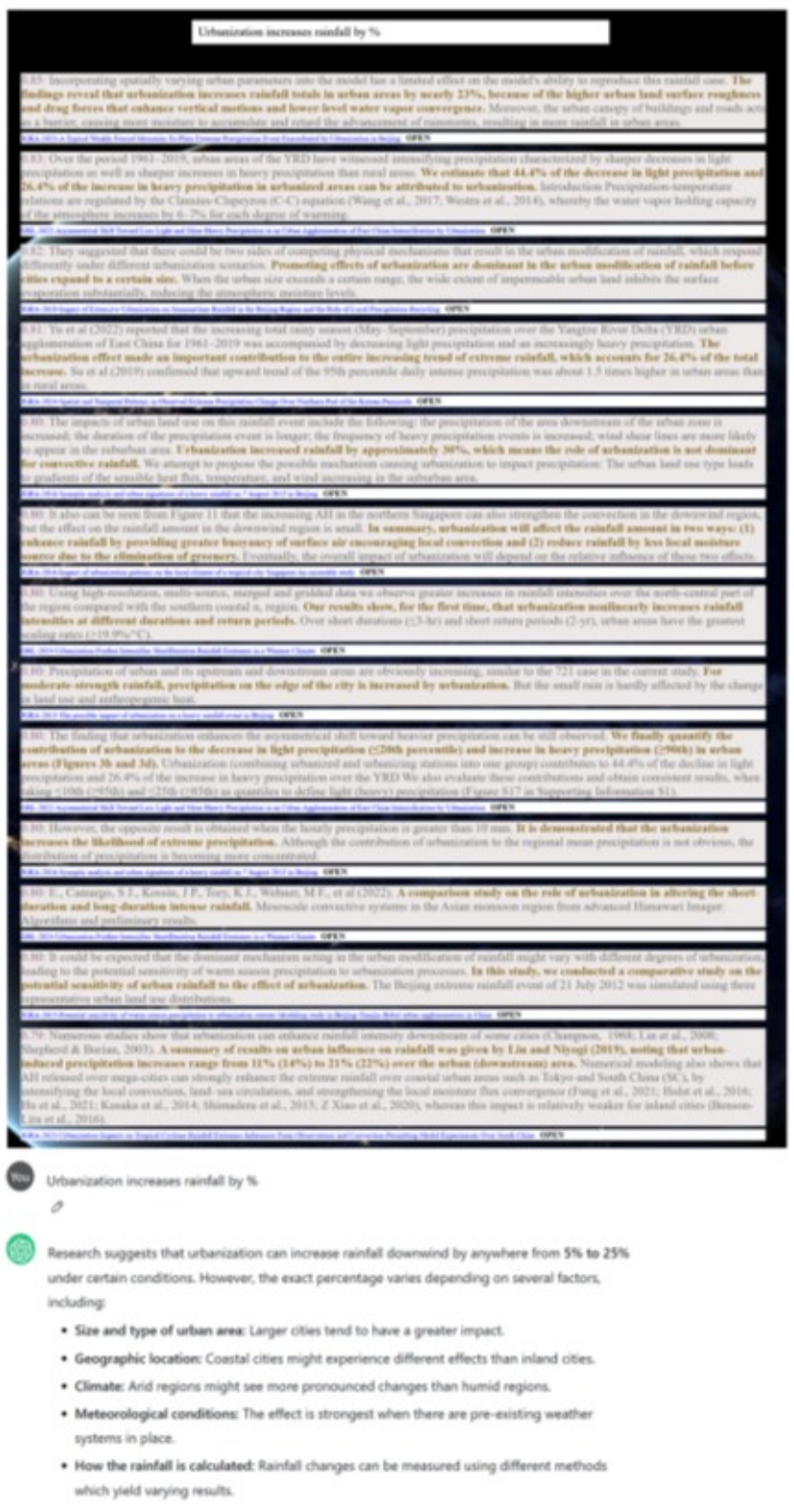

**Figure S5.** The query "Urbanization increases rainfall by %" was submitted. The output from software was demonstrably clearer than that of ChatGPT-4.

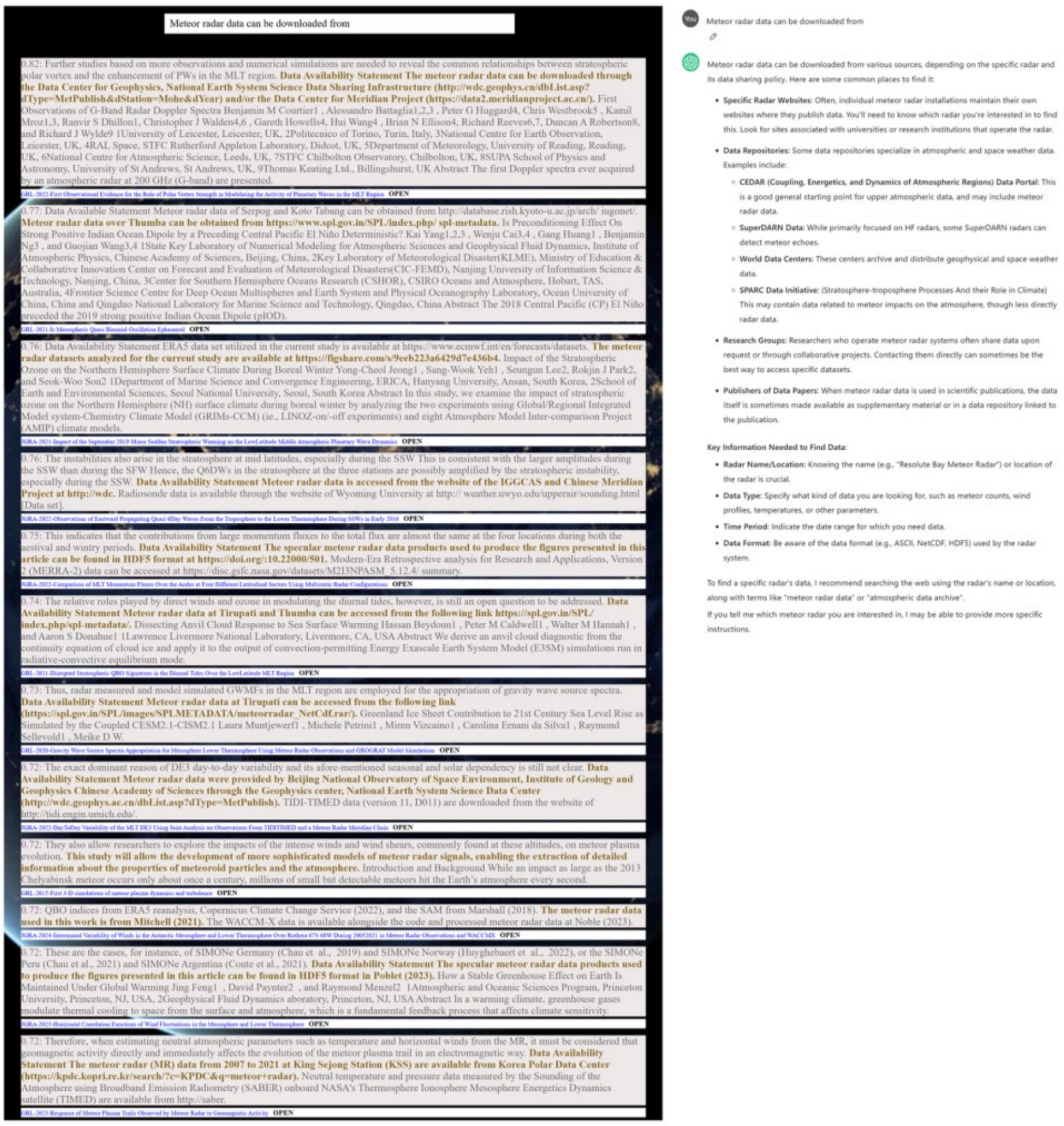

**Figure S6.** The query "Meteor radar data can be downloaded from" was submitted. The software provided numerous website suggestions. In contrast, ChatGPT-4 was unable to provide an accurate download website address.

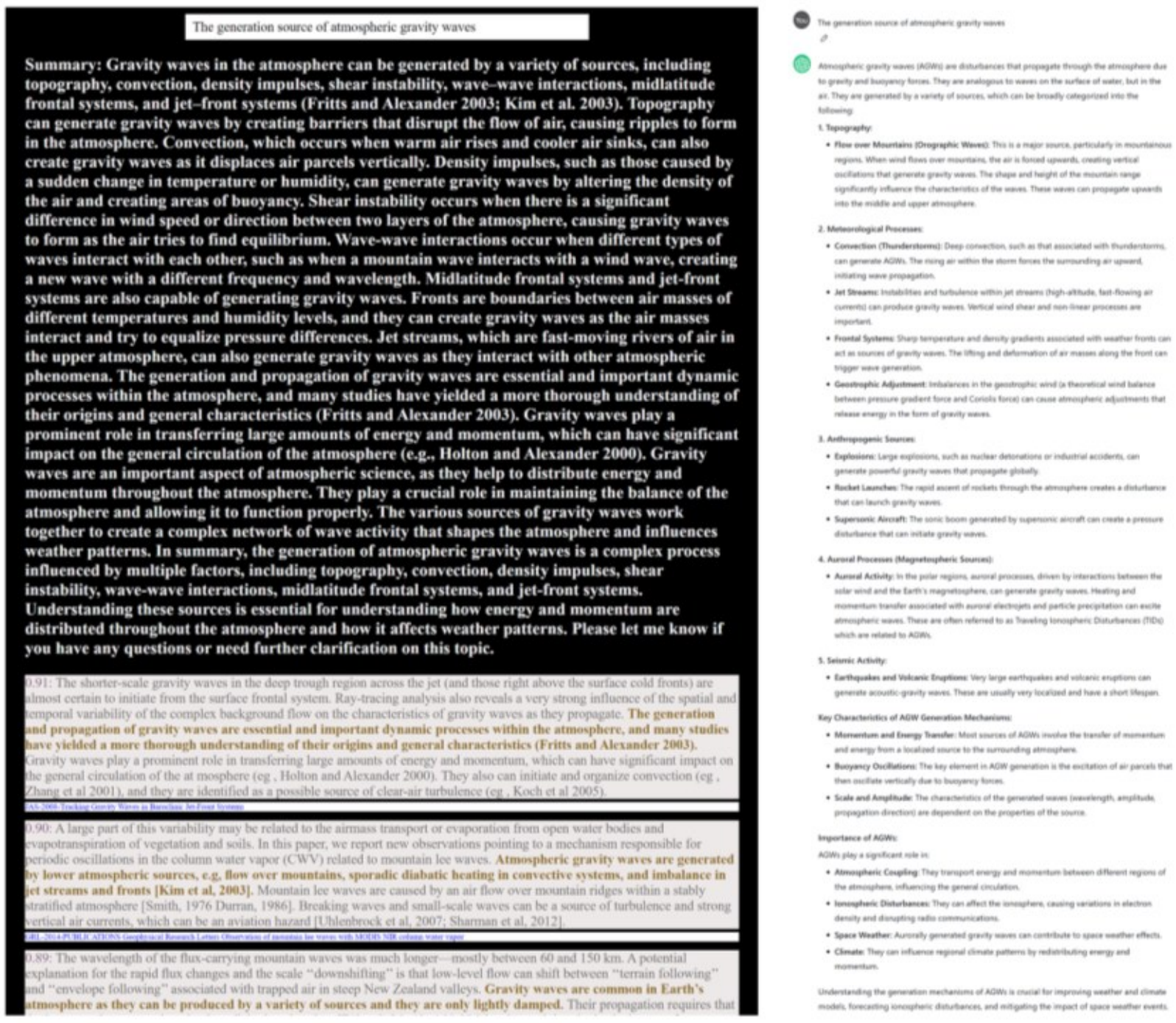

**Figure S7.** The query "The generation source of atmospheric gravity waves" was submitted. The bold, white text represents the summary generated by an LLM (Llama 2 70B). This summary correctly references the work of Fritts & Alexander (2003), a seminal publication in the field of gravity waves. In contrast, outputs from ChatGPT-4 failed to elucidate the full mechanism of gravity wave generation, including processes such as wind shear and wave-wave interactions.

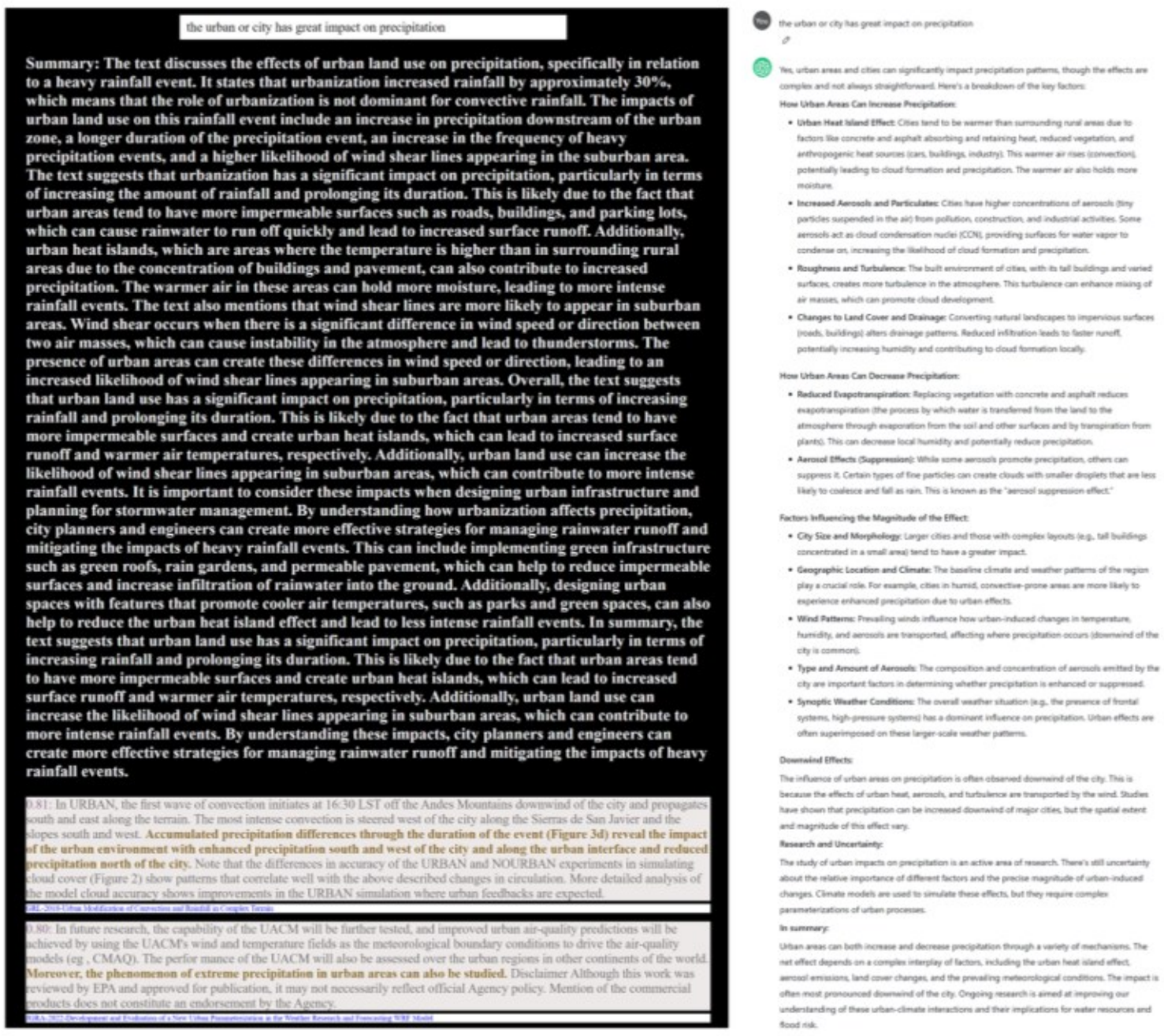

**Figure S8.** The query "the urban or city has great impact on precipitation" was submitted. The LLM summary (Llama 2 70B) can accurately capture the data presented in Fig. S5.

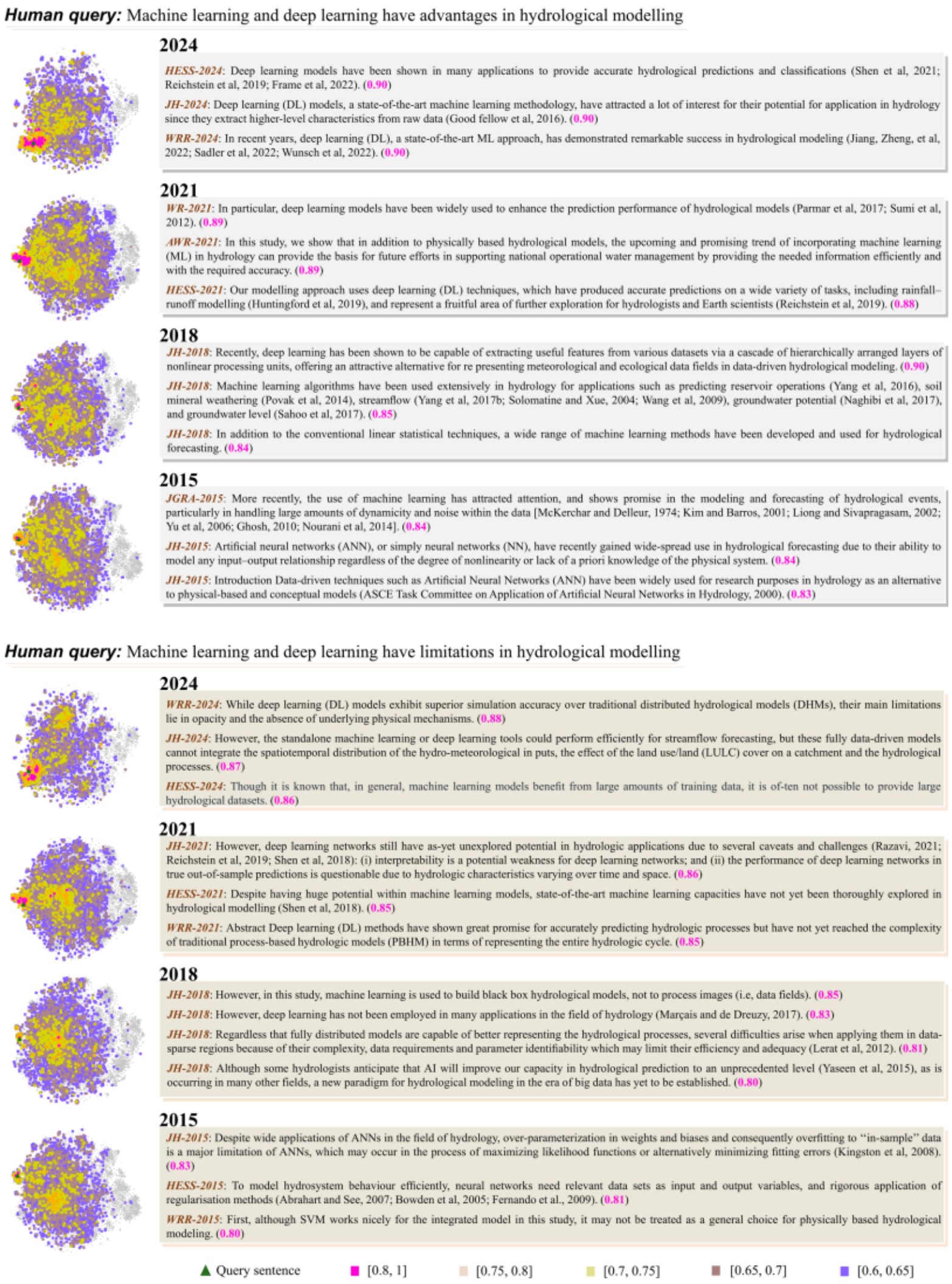

**Figure S9.** Taking "machine learning and deep learning in hydrological modelling" as an illustrative example to explore semantic search across two semantically similar tasks. The human queries are "Machine learning and deep learning have advantages in hydrological modelling" (above) and "Machine learning and deep learning have limitations in hydrological modelling" (down). Sentence libraries from the years 2024, 2021, 2018, and 2015, containing the keywords "hydro" (case-insensitive) and "model" (case-insensitive) are utilized. The total number of sentences ranges from 8,898 in 2015 to 22,155 in 2021. All sentences are transformed into 1024-dimensional dense vectors using PSTM_4. The top-ranked sentences for each year are displayed in the colored box.

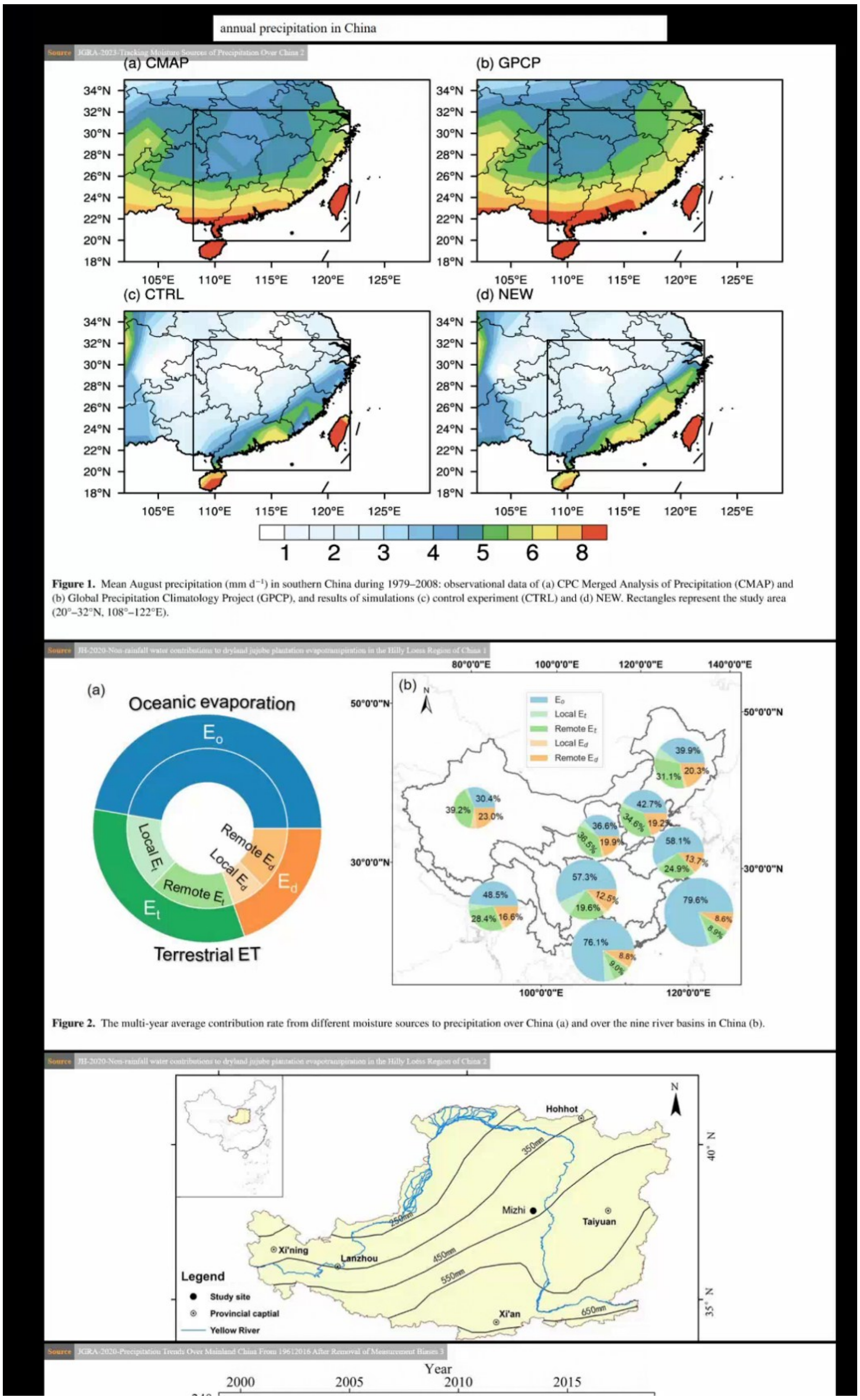

**Figure S10.** Professional image retrieval software can identify related images by performing semantic matching between user-provided prompts and academic paper image captions. This approach enables students, teachers or researchers to efficiently locate professional images with a high resolution. For instance, the prompt “annual precipitation in China” was submitted.

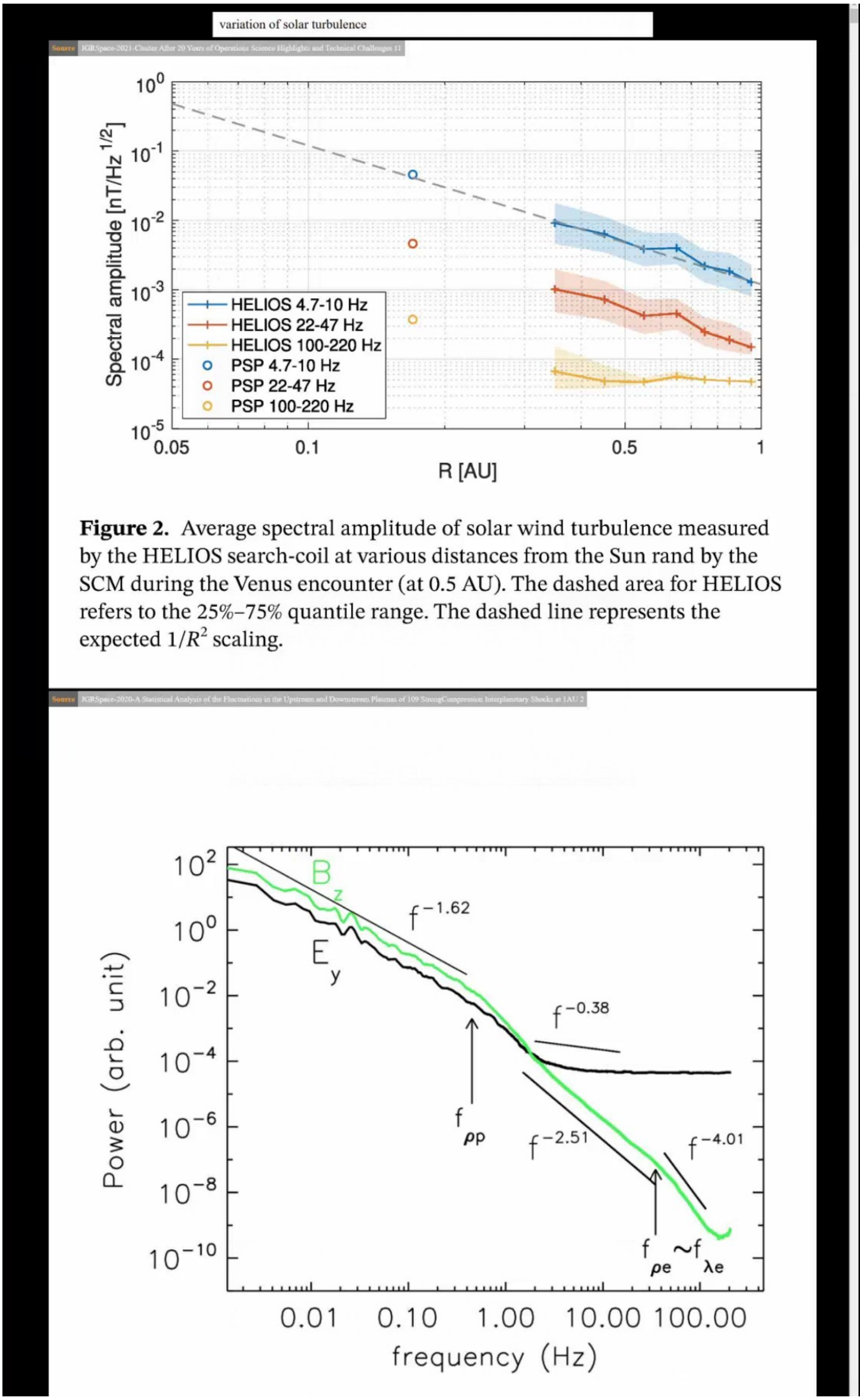

**Figure S11.** Similar to Fig. S10, but for the prompt of “variation of solar turbulence”.